\documentclass[letterpaper, 10 pt, conference]{ieeeconf}  

\IEEEoverridecommandlockouts                              

\usepackage{graphicx} 
\usepackage{epsfig} 
\usepackage{mathptmx} 
\usepackage{times} 
\usepackage{amsmath} 
\usepackage{amssymb}  
\usepackage{caption}
\usepackage{subcaption}
\usepackage{multirow}
\usepackage{comment}
\usepackage{hyperref}
\usepackage{xcolor}
\usepackage{color}
\usepackage{soul}
\sethlcolor{yellow}

\newcommand{\change}[1]{#1}

\newcommand{\changedel}[1]{}

\newcommand{\andal}[1]{#1 et al.}
\newcommand{\tabcell}[2][c]{%
  \begin{tabular}[#1]{@{}c@{}}#2\end{tabular}}

\title{\LARGE \bf
ParisLuco3D: A high-quality target dataset for domain generalization of LiDAR perception}%
\author{Jules Sanchez$^{1}$*, Louis Soum-Fontez$^{1}$*, Jean-Emmanuel Deschaud$^{1}$, and Francois Goulette$^{1,}$$^{2}$
\thanks{*Equal contributions}
\thanks{$^{1}$Centre for Robotics, Mines Paris - PSL, 
PSL University, 
75006 Paris, France
{\tt\scriptsize firstname.surname@minesparis.psl.eu}}%
\thanks{$^{2}$U2IS, ENSTA Paris, Institut Polytechnique de Paris, 91120 Palaiseau, France
{\tt\scriptsize firstname.surname@ensta-paris.fr}}%
}

\begin{document}

\maketitle
\thispagestyle{empty}
\pagestyle{empty}

\begin{abstract}

LiDAR is an essential sensor for autonomous driving by collecting precise geometric information regarding a scene. 
As the performance of various LiDAR perception tasks has improved, generalizations to new environments and sensors has emerged to test these optimized models in real-world conditions. 
Unfortunately, the various annotation strategies of data providers complicate the computation of cross-domain performances.

This paper provides a novel dataset, ParisLuco3D, specifically designed for cross-domain evaluation to make it easier to evaluate the performance utilizing various source datasets. Alongside the dataset, online benchmarks for LiDAR semantic segmentation, LiDAR object detection, and LiDAR tracking are provided to ensure a fair comparison across methods. 

The ParisLuco3D dataset, evaluation scripts, and links to benchmarks can be found at the following website:\\
\url{https://npm3d.fr/parisluco3d}

\end{abstract}

\section{Introduction}

LiDAR-based perception for autonomous driving applications has become increasingly popular in the last few years. LiDAR provides reliable and precise geometric information and is a useful addition to typical camera-based systems.
The various LiDAR perception tasks have gained access to a growing number of open-source datasets. This large amount of data, and the thorough efforts of the community, have led to very good performance, notably by Cylinder3D \cite{cylinder3d} for LiDAR Semantic Segmentation (LSS) and CenterPoint \cite{Yin_2021_CVPR} for LiDAR Object Detection (LOD).

Consequently, new tasks for evaluating the robustness of these methods in new scenarios and environments have emerged. Here, we focus on methods that deal with domain generalization, which have become an increasing focus in LiDAR perception. These methods involve confronting a model trained on a specific domain, with a new domain at the time of inference. In practice, a subset of datasets are used for training, and other datasets expected to be acquired elsewhere are used for testing. 

While open-source datasets display a large variety of scenes and sensor setups, crucial discrepancies have been identified among label sets \cite{Kim_2023_CVPR,sanchez2023cola}. \change{Moreover, despite previous works managing to provide meaningful insights in the domain generalization task, the lack of consensual labels sets can result in unfair comparisons.}
\changedel{Due to this label difference, it is not possible to leverage current online benchmarks to ensure fair comparisons across methods.}


\begin{figure}
    \centering

    \includegraphics[trim={7cm 7cm 10cm 5cm},clip,width=0.85\linewidth]{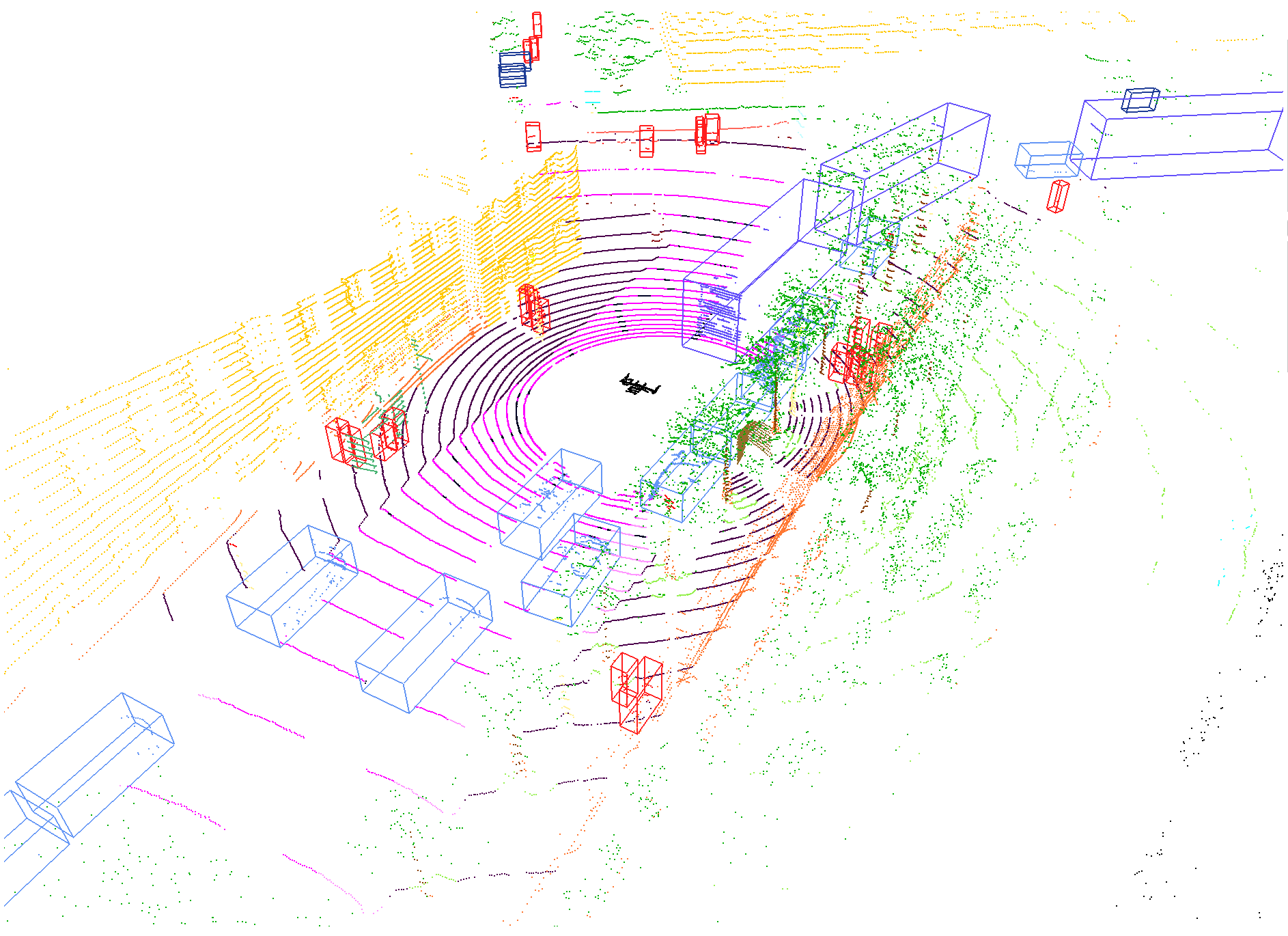}

    \caption{A LiDAR scan of our ParisLuco3D dataset with ground truth annotation for semantic segmentation and object detection.}
    \label{fig:extract}
\end{figure}

\begin{table*}[h]
\scriptsize
    \centering
    \begin{tabular}{c|c|c|c|c|c|c|c|c|c|c|c}
    \multirow{2}{*}{Name} &\multirow{2}{*}{Year}& \multirow{2}{*}{\tabcell{\# scans\\ train for LSS}} & \multirow{2}{*}{\tabcell{\# scans\\ validation for LSS}} & \multirow{2}{*}{LSS} & \multirow{2}{*}{LOD}& \multirow{2}{*}{Tracking} & \multirow{2}{*}{Location} & \multirow{2}{*}{\tabcell{Scene \\ type}}  &\multicolumn{2}{c|}{\tabcell{Sensor resolution ($^{\circ}$)}}&  \multirow{2}{*}{\# beams}\\ \cline{10-11}
          & & &  &  &  &  &  & & \tabcell{vertical } & \tabcell{horizont.}\\\hline\hline
       KITTI \cite{Geiger2012CVPR}& 2012&N/A&N/A& $\times$ & $\sim$  & $\times$ & \tabcell{Germany} & \tabcell{suburban} & 0.4& 0.08  & 64\\\hline
       SemanticKITTI \cite{behley2019iccv}&2019 & 19k &4071& \checkmark & $\times$  & \checkmark & \tabcell{Germany} & suburban & 0.4 & 0.08 & 64 \\\hline
       nuScenes \cite{nuscenes}&2020 & 29k &6019 & \checkmark & \checkmark & \checkmark  & \tabcell{US+Singapore} & urban & 1.33 & 0.32 & 32 \\\hline
       SemanticPOSS \cite{semanticposs}& 2020&N/A&2988$^\dagger$& \checkmark & $\times$  & $\times$  & \tabcell{China} & campus & 0.33-6 & 0.2 & 40 \\\hline
       Waymo \cite{waymo}& 2020& 24k &5976 & \checkmark & \checkmark  & \checkmark & \tabcell{US} & urban & 0.31& 0.16 & 64 \\\hline
       ONCE \cite{once_ds}&2021 &N/A&N/A&$\times$ & $\checkmark$  & $\times$ &  \tabcell{China} & \tabcell{sub.+urban} & 0.33-6 & 0.2 & 40 \\\hline
       Pandaset \cite{pandaset}&2021 &N/A&6080$^\dagger$ & \checkmark & \checkmark  & \checkmark & \tabcell{US} & suburban & 0.17-5 & 0.2 & 64\\\hline
       HelixNet \cite{helix4D}& 2022&49k&8179  & \checkmark& $\times$ &  $\times$ & \tabcell{France} & \tabcell{sub.+urban} & 0.4& 0.08 & 64 \\\hline
       KITTI-360 \cite{kitti360}&2023 &60k&15204 & $\sim$ & \checkmark  & $\times$ & \tabcell{Germany} & suburban & 0.4 & 0.08 & 64 \\\hline\hline
        ParisLuco3D (Ours) & 2023 &N/A& 7501$^\dagger$ & \checkmark & \checkmark & \checkmark & France & urban & 1.33 & 0.16 & 32\\
    \end{tabular}
    \caption{Summary of the various existing LiDAR perception datasets. $^\dagger$size of the full dataset, as there is no designated split. KITTI provides LOD annotation only for a subset of the point cloud. KITTI-360 provides the LSS annotation on the accumulated and subsampled point cloud rather than on each scan (LSS = LiDAR Semantic Segmentation and LOD = LiDAR Object Detection).}
    \label{tab:recap_data}
\end{table*}

In this paper, we propose a new dataset that is specifically designed for domain generalization evaluation, with annotations that have been mapped to existing standard datasets to avoid the need for comprise in quality evaluations of the domain generalization methods. Since this dataset is focused only on cross-domain generalization evaluation, its annotations are not made available to the public.

We make the following contributions: 

\begin{itemize}
    \item The release of a novel dataset, in addition to online benchmarks, to compute cross-domain performances fairly. \change{This dataset was acquired in the center of Paris, a relatively unavailable scene type.} 
    \item A thorough overview of the generalization capabilities of current state-of-the art architecture for both semantic segmentation and object detection in order to provide the community with a baseline for these tasks.
\end{itemize}

\section{Related Work}
\subsection{LiDAR Datasets in autonomous driving}

The tasks that we focus on for LiDAR perception are LiDAR Semantic Segmentation (LSS), LiDAR Object Detection (LOD) and Tracking.

A summary of the main datasets along with the tasks they are annotated for, sensor information, and the amount of scans available is presented in \autoref{tab:recap_data}. 

The two most common perception tasks, measured by the number of annotations available or the number of published works in these fields, are semantic segmentation and object detection. Accordingly, these tasks are the focus of the remainder of this work. Recently, some large scale datasets have been released, such as HelixNet \cite{helix4D}, DurLAR \cite{2021durlar3dv} or Argoverse 2 \cite{wilson2023argoverse} but due to their recency or lack of annotations, they are not yet widely used in the LSS and LOD literature.

\subsubsection{Datasets for LiDAR semantic segmentation}

The first dataset released for LSS was SemanticKITTI~\cite{behley2019iccv} in 2019, which was derived from the KITTI \cite{Geiger2012CVPR} dataset. It was acquired in a German suburb. Due to its size and seniority, it is the reference dataset for LiDAR semantic segmentation. Its annotations are evenly split between road users and background classes.

Thereafter, SemanticKITTI was expanded into KITTI-360 \cite{kitti360}, which includes data on the same suburb but on a larger scale.
Subsequently, a few object detection datasets were expanded to incorporate semantic segmentation annotation, such as nuScenes \cite{nuscenes} and Waymo \cite{waymo}. They were acquired in cities and suburbs in the US and Asia. Contrary to SemanticKITTI, nuScenes focuses its annotations on vehicles with a limited number of background classes.

Finally, a few additional targeted datasets were released, such as SemanticPOSS \cite{semanticposs}, which proposed a new setup by providing a dataset acquired in a student campus, and Pandaset \cite{pandaset}, which has a double-sensor setup.


\subsubsection{Datasets for LiDAR object detection}

In 2014, KITTI~\cite{Geiger2012CVPR} was released and ushered in a new range of experimental data for LOD. 
While there are several downsides in its annotation, such as the low number of classes and the limited annotated field of view, it has remained an important benchmark for detection due to the quality of its annotations and its high-density point clouds. 

The nuScenes \cite{nuscenes} dataset was subsequently released with full 360-degrees annotations and an exhaustive list of annotated classes. Even though the number of available scans in this dataset is higher than those in KITTI, they are much sparser due to the frequency of the sensor and its resolution. 

The Waymo \cite{waymo} dataset is the one with the most 3D annotations, along with a high number of high-density point clouds. Nevertheless, it only includes coarse class information, whereas other datasets have a finer label space. 

Finally, new datasets were released in an attempt to diversify the environments and LiDAR sensors used for detection, such as the ONCE \cite{once_ds} dataset. This dataset uses a 40-beam LiDAR sensor, therby resulting in an atypical point cloud density distribution. 

\subsection{Domain generalization for LiDAR perception}

Domain generalization was long reserved for 2D perception \cite{generalization}, as domain adaptation was the main focus of 3D research. The difference between these fields is that domain generalization does not utilize any information before processing the new domain. But models have been found to be very sensitive to domain gaps in datasets such as differences in LiDAR density or object sizes \cite{wang2020train}.

Recently, a few studies on domain generalization of LiDAR perception have emerged, particularly for object detection \cite{lehner20223d} by adversarial augmentation and for semantic segmentation \cite{Kim_2023_CVPR,3DLabelProp,completelabel} using domain alignment by identifying a canonical representation either in the feature space or in the Euclidean space. Additionally, MDT3D \cite{soumfontez2023mdt3d} is focused on multi-source domain generalization. 

All the recent studies highlight the difficulties of performing cross-domain evaluation, as annotations can and do differ from one dataset to another. Specifically, they have to resort to studying either the intersection of the label set \cite{completelabel}, usually on very few labels, or a remapping to a common and coarser label set \cite{Kim_2023_CVPR}. 

ParisLuco3D provides a new annotation set that has been mapped to standard LiDAR perception datasets. Cross-domain evaluation can be performed fairly without any compromise regarding the fineness of the label set. Furthermore, while this dataset could be considered small, as it is only used for evaluation, it is on par or even bigger than most validation sets (\autoref{tab:recap_data}).

\section{Dataset presentation}

\subsection{Acquisition and data generation}

\subsubsection{Acquisition}

\begin{figure}[h]
    \centering
    \includegraphics[width=0.6\linewidth]{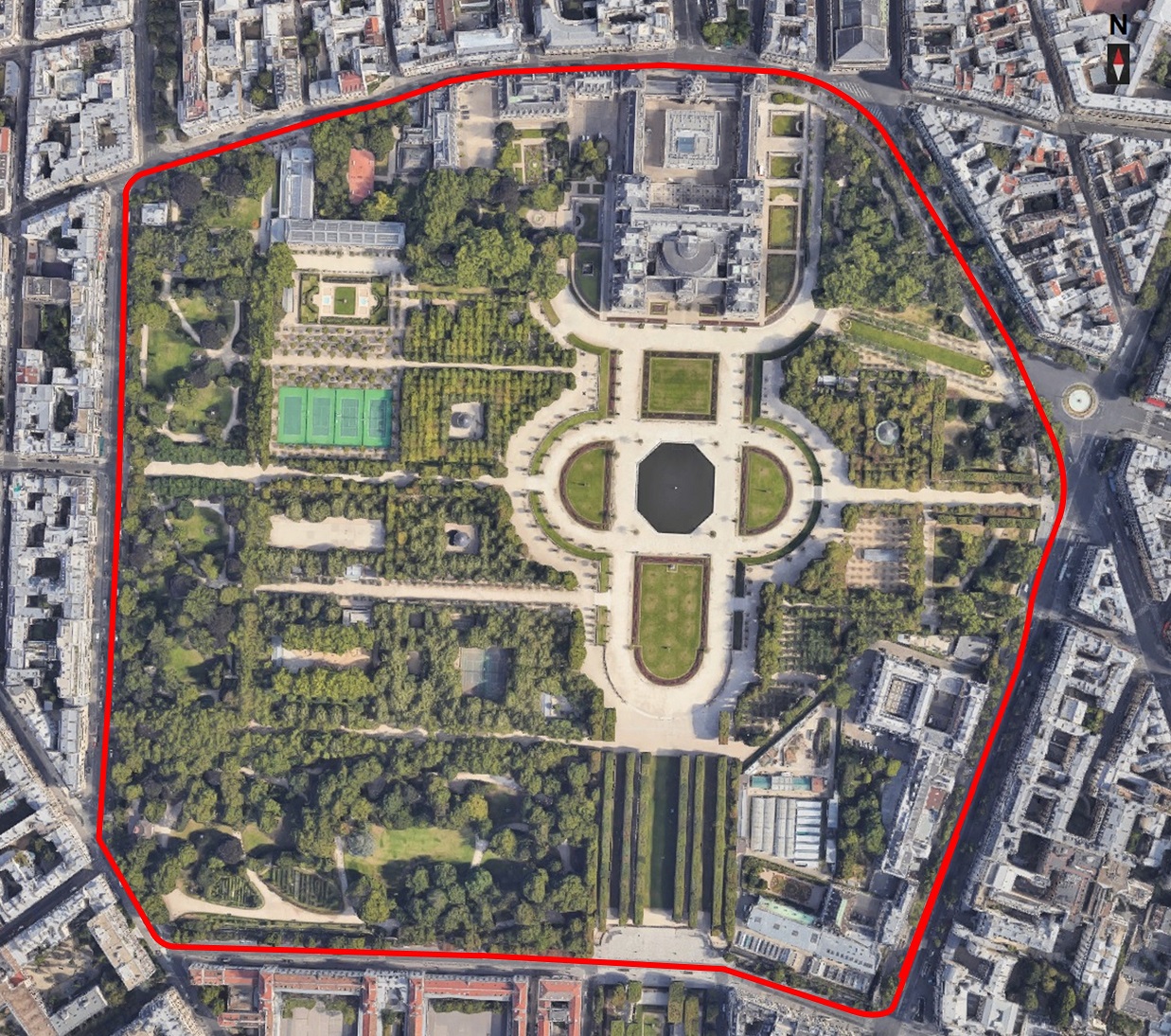}
    \caption{Trajectory of our dataset (2.1~\text{km} around the Luxembourg Garden in Paris) overlaid on a Google Satellite Image.}
    \label{fig:traj}
\end{figure}

\begin{figure}
     \centering
     \begin{subfigure}[b]{0.49\linewidth}
         \centering
         \includegraphics[width=\linewidth]{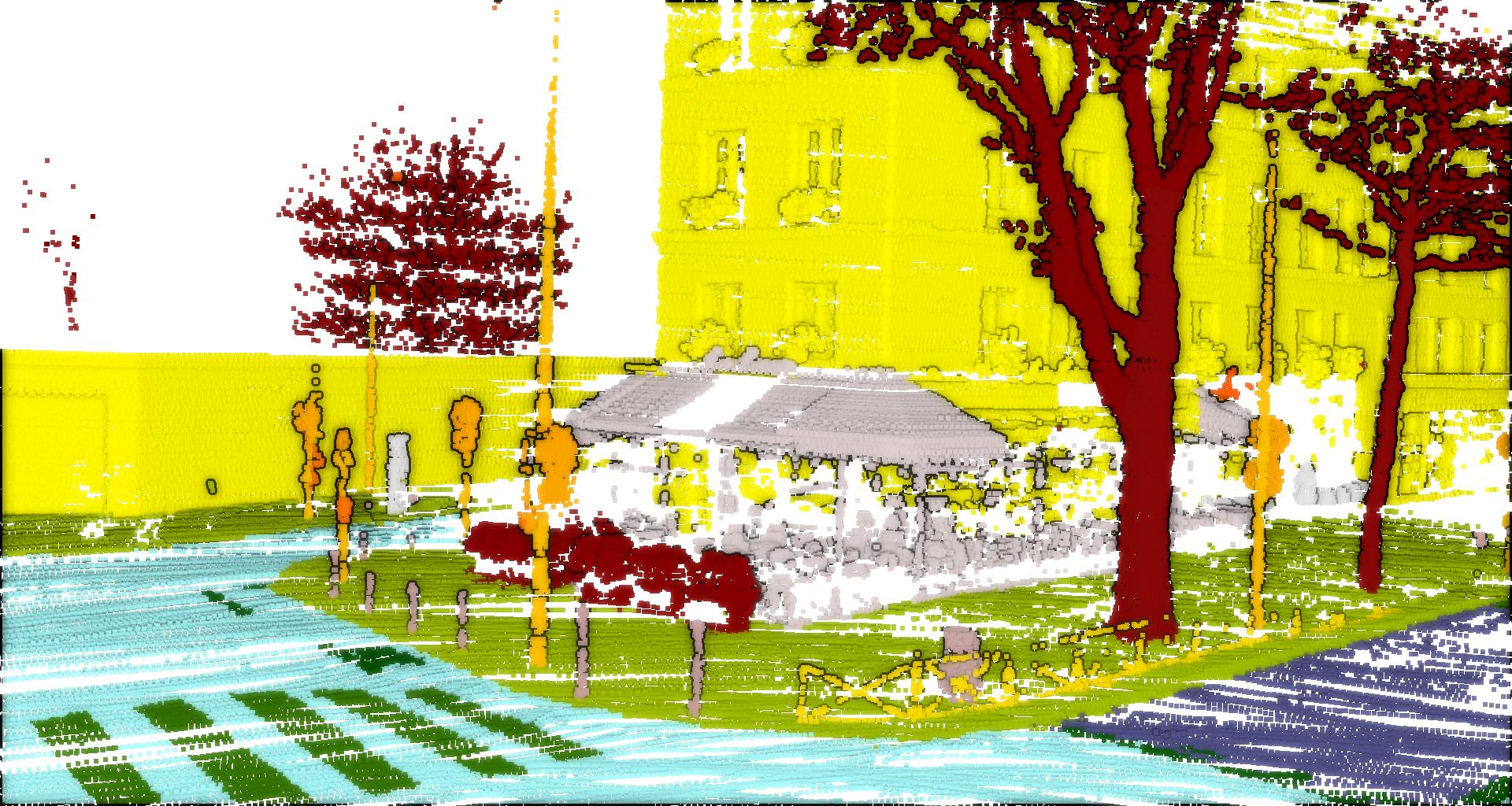}
     \end{subfigure}
     \begin{subfigure}[b]{0.49\linewidth}
         \centering
         \includegraphics[width=\linewidth]{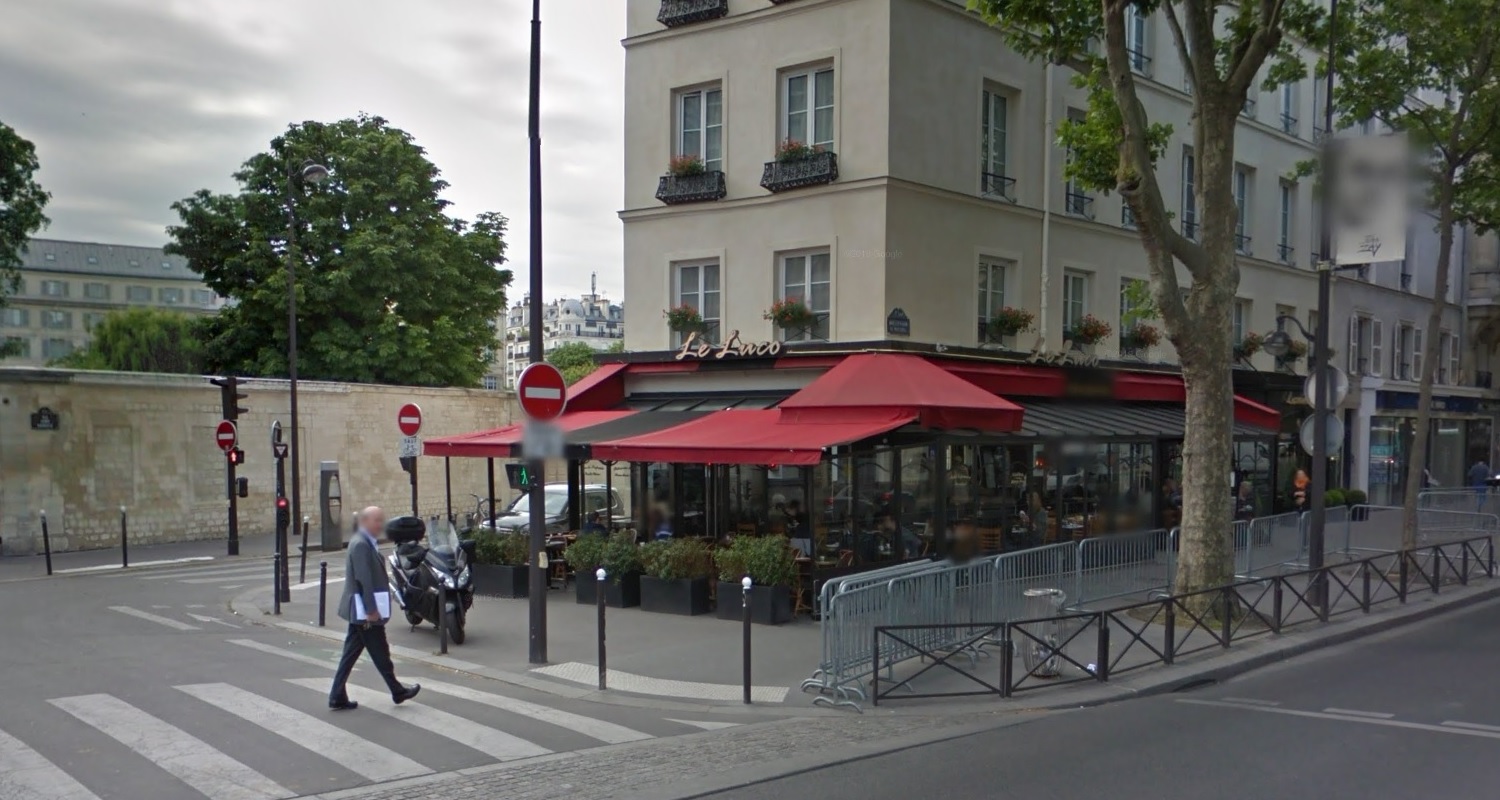}
     \end{subfigure}
     \begin{subfigure}[b]{0.49\linewidth}
         \centering
         \includegraphics[width=\linewidth]{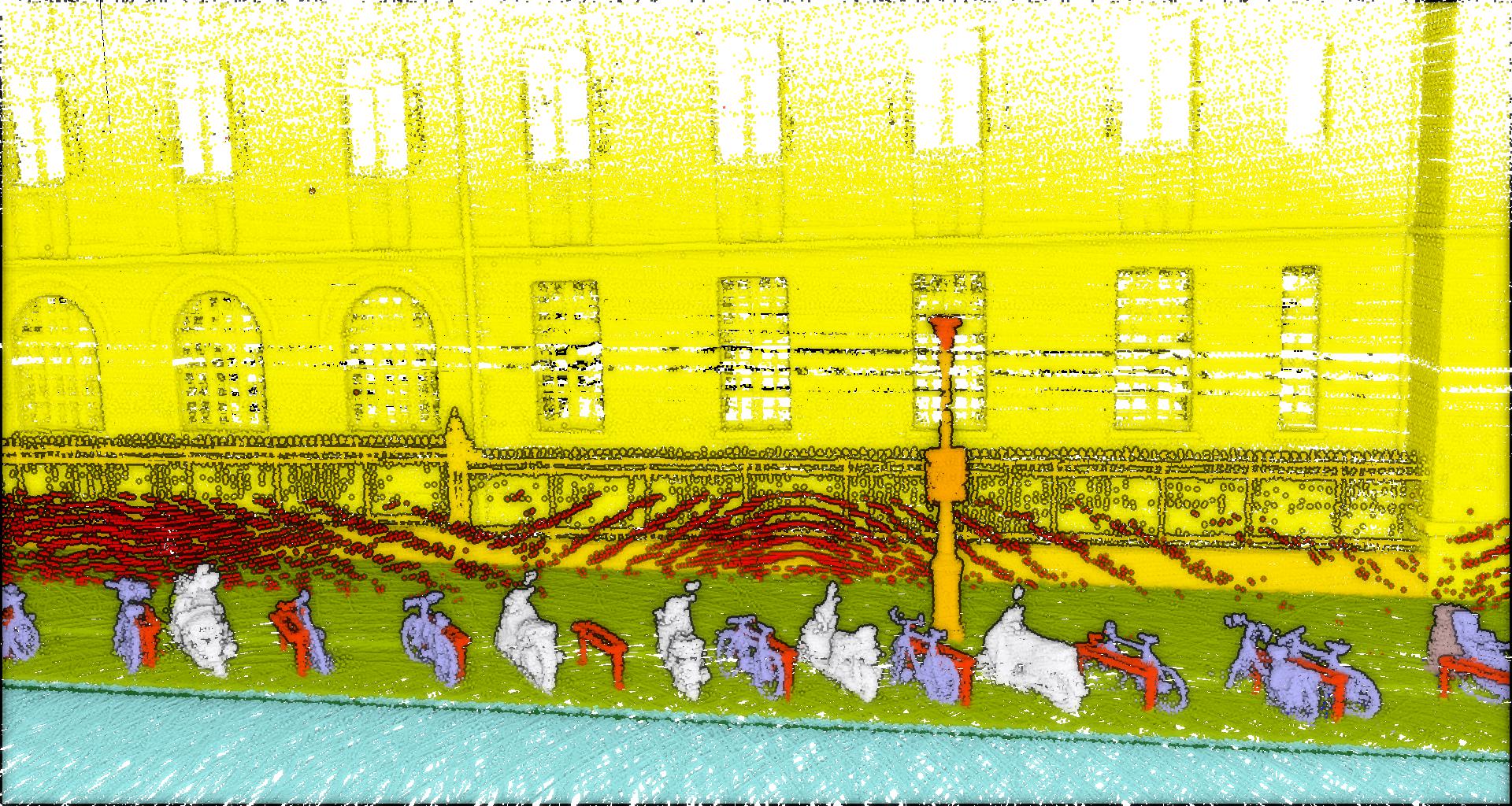}
     \end{subfigure}
     \begin{subfigure}[b]{0.49\linewidth}
         \centering
         \includegraphics[width=\linewidth]{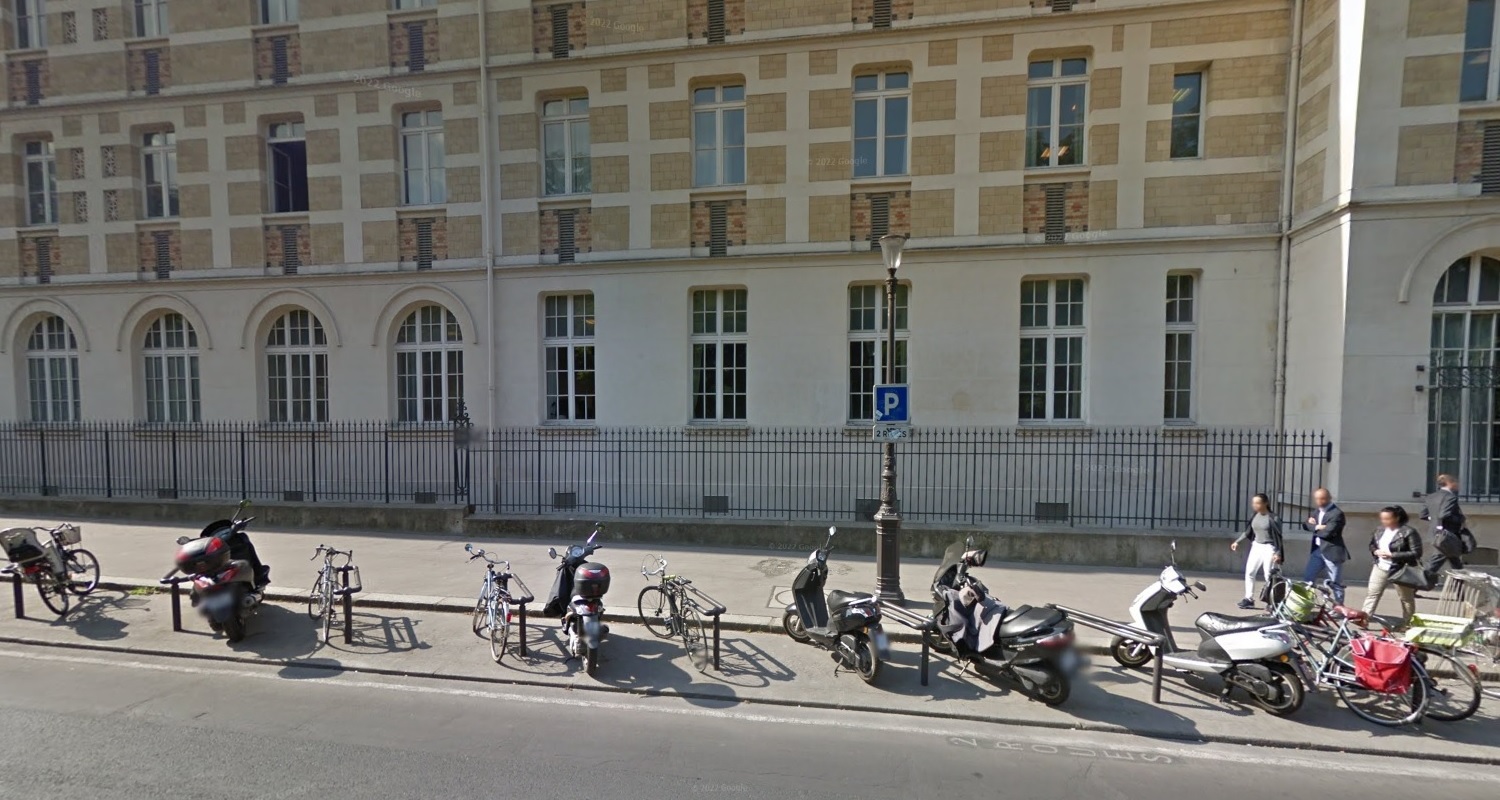}
     \end{subfigure}
        \caption{Illustrations depicting the diversity of scenes and quality of annotation in the ParisLuco3D dataset (left: point clouds by accumulating the scans of our dataset colorized with labels\protect\footnotemark; right: images taken from Google Street View). }
        \label{fig:illustrations_streetview}
\end{figure}

The acquisition of the ParisLuco3D dataset was performed in Paris in November 2015 over a distance of 2.1~\text{km}. The trajectory, depicted in Figure~\ref{fig:traj}, is around the Luxembourg Garden in the heart of the city of Paris.
Figure~\ref{fig:illustrations_streetview} depicts point clouds by accumulating the scans and illustrates the diversity of scenes and contexts in our dataset.

The LiDAR sensor used was a Velodyne HDL32 positioned vertically on a pole that was attached to the roof of a Citröen Jumper vehicle. Due to its placement and size of the vehicle, the sensor was approximately 3.70 m from the ground (a height that is different from what is usually available in other LiDAR datasets in autonomous driving).



\subsubsection{Data generation}

The ParisLuco3D dataset consists of 7501 scans in the form of 3D point clouds (a scan is a 360-degree horizontal rotation of the LiDAR sensor rotating at 10Hz). The data are available in raw format with the following information: \textit{x}, \textit{y}, \textit{z}, \textit{timestamp}, \textit{intensity}, and \textit{laser\_index}.
\textit{Timestamp} is the time per point provided by the LiDAR sensor, synchronized with a GPS sensor.
\textit{Intensity} is the strength of the return signal, which represents 256 calibrated reflectivity values (diffuse reflectors are represented with values ranging from $0$ to $100$ and retro-reflectors from 101\textendash255)\footnote{\url{https://tinyurl.com/ycxm2d7a}}.
\textit{Laser\_index} refers to the number of the laser that was fired (32 lasers in the Velodyne HDL32). \textit{Timestamp} and \textit{laser\_index}, which are often not available in other datasets~\cite{behley2019iccv,nuscenes,waymo,semanticposs,pandaset}, can enable the testing of specific methods, such as the Helix4D~\cite{helix4D}, which exploits the timestamp.

This raw data do not take into account the movement of the vehicle during a scan. To estimate this movement, we used CT-ICP~\cite{cticp} LiDAR odometry, which enables the calculation of the deformation of the scans. Therefore, we also provide motion-corrected scans with the same information as raw scans: \textit{x}, \textit{y}, \textit{z}, \textit{timestamp}, \textit{intensity}, and \textit{laser\_index}. All data annotations (semantic class per points, bounding boxes of objects, and tracking of objects) were made on these motion-corrected scans. 

Finally, to enable the use of methods that utilize data sequentiality (i.e., past scans), we again used CT-ICP to compute a pose for each scan (motion-corrected) in a world frame corresponding to the frame of the first scan of the sequence. The poses are provided in KITTI format~\cite{Geiger2012CVPR} and correspond precisely to the pose of the vehicle in the middle of a scan acquisition (i.e., when the LiDAR lasers fire forward like the KITTI convention).

The sequence is composed of a loop, with an overlap of approximately 150~\text{m} between the start and end of the acquisition. A registration by ICP on the overlap zone and a graph SLAM solver~\cite{graphslam} enabled us to calculate the final poses with loop closure (Figure~\ref{fig:traj}).

\subsection{Label shift and annotation process}

\subsubsection{Label shift}

A major issue when conducting cross-dataset evaluation is the difference in label set definitions. There are three sources of differences, which we illustrate for the semantic segmentation case. 

First, certain objects are annotated in one dataset but not in another. An example of this is the \textit{traffic-cone} annotation found in nuScenes but not in SemanticKITTI. 

Second, there are granularity differences. For example, nuScenes defines \textit{manmade}, whereas SemanticKITTI distinguishes \textit{building}, \textit{pole}, and \textit{sign}. Similarly, Waymo uses a single \textit{Vehicle} class instead of making a distinction between cars, trucks, and buses.

Finally, there is a phenomenon we call the \textit{label shift}, which represents different objects under similar labels. An example of label shift is the definitions of \textit{road} and \textit{sidewalk} in nuScenes and SemanticKITTI. In one dataset, bike lanes are part of the \textit{road} label, whereas in the other dataset they part of the \textit{sidewalk} label.


Furthermore, \textit{label shift} appears and results in change for bounding box dimensions. For example, the Waymo dataset does not include large objects carried by pedestrians in its bounding boxes and does not annotate any parked bicycles.

\subsubsection{Annotation process}

In order to guarantee the quality of the annotations in this dataset, we relied on expert annotators working in the field of 3D point cloud processing who have experience in working with existing datasets. The dataset was triple-checked in order to identify any errors.


Semantic annotation was done on the dense point clouds accumulated from motion-corrected scans. The annotation of objects for detection and tracking was done scan by scan by exploiting neighboring scans when the objects were occluded with a bounding box interpolated over time. The software used for the semantic segmentation annotation was \textit{point labeler} \cite{behley2019iccv}, while the one for object detection was \textit{labelCloud} \cite{labelcloud}, which are both open-source. 

%

\subsection{Labels for semantic segmentation}

\begin{table}[h]
    \centering
    \scriptsize
    \begin{tabular}{c|c||c|c|c}
        Dataset & \#labels & \#labels$\cap$SK &\#labels$\cap$NS&\#labels$\cap$PL\\\hline\hline
        SemanticKITTI \cite{behley2019iccv}&19 & N/A & 10 &\textbf{19}\\ \hline
        nuScenes \cite{nuscenes} &16 & 10 & N/A &\textbf{16}\\ \hline
        Pandaset \cite{pandaset}&37& 8 &8 &\textbf{13}\\ \hline
        Waymo \cite{waymo}&22 & 15 & 11 &\textbf{20}\\ \hline
        KITTI-360 \cite{kitti360}&18 & 15 & 13 &\textbf{16}\\ \hline
        SemanticPOSS \cite{semanticposs}&13 &11 & 6 &\textbf{13}\\ \hline
        HelixNet \cite{helix4D}& 9 & 7 & 6 & \textbf{7}\\ \hline \hline
        ParisLuco3D & 45 & 19 & 16 & N/A
        
    \end{tabular}
    \caption{Number of labels of each LSS dataset and the size of the intersection of their label set with SemanticKITTI (SK) and nuScenes (NS).}
    \label{tab:labels}
\end{table}

In \autoref{tab:labels}, we summarize the number of labels from the various datasets and indicate how many of these labels are intersecting with the label sets of nuScenes and SemanticKITTI. We chose these two datasets because they are the commonly used training sets. When cross-domain evaluation is performed, we see a significant reduction in the number of labels for evaluation and a loss of fineness for prediction.

For ParisLuco3D, we wanted to define a label set that could be easily mapped to nuScenes and SemanticKITTI, as they are the two reference training sets. As such, the annotation details of nuScenes and SemanticKITTI were dissected to understand which objects are encapsulated in each label to avoid a label shift between ParisLuco3D and these datasets. This results in fine annotation, thereby avoiding any ambiguity in the remapping.

There are then 45 labels for semantic segmentation: \textit{car}, \textit{bicycle}, \textit{bicyclist}, \textit{bus}, \textit{motorcycle}, \textit{motorcyclist}, \textit{scooter}, \textit{truck}, \textit{construction-vehicle}, \textit{trailer}, \textit{person}, \textit{road}, \textit{bus-lane}, \textit{bike-lane}, \textit{parking}, \textit{road-marking}, \textit{zebra-crosswalk}, \textit{roundabout}, \textit{sidewalk}, \textit{central-median}, \textit{building}, \textit{fence}, \textit{pole}, \textit{traffic-sign}, \textit{bus-stop}, \textit{traffic-light}, \textit{light-pole}, \textit{bike-rack}, \textit{parking-entrance}, \textit{metro-entrance}, \textit{vegetation}, \textit{trunk}, \textit{vegetation-fence}, \textit{terrain}, \textit{temporary-barrier}, \textit{pedestrian-post}, \textit{garbage-can}, \textit{garbage-container}, \textit{bike-post}, \textit{bench}, \textit{ad-spot}, \textit{restaurant-terrace}, \textit{road-post}, \textit{traffic-cone}, and \textit{other-object}. 

These labels cover the usual autonomous driving labels as well as the more specific features of specific cases of the Paris landscape, such as metro entrances or bar terraces. 

\begin{figure}[h]
    \centering
    \includegraphics[width=\linewidth]{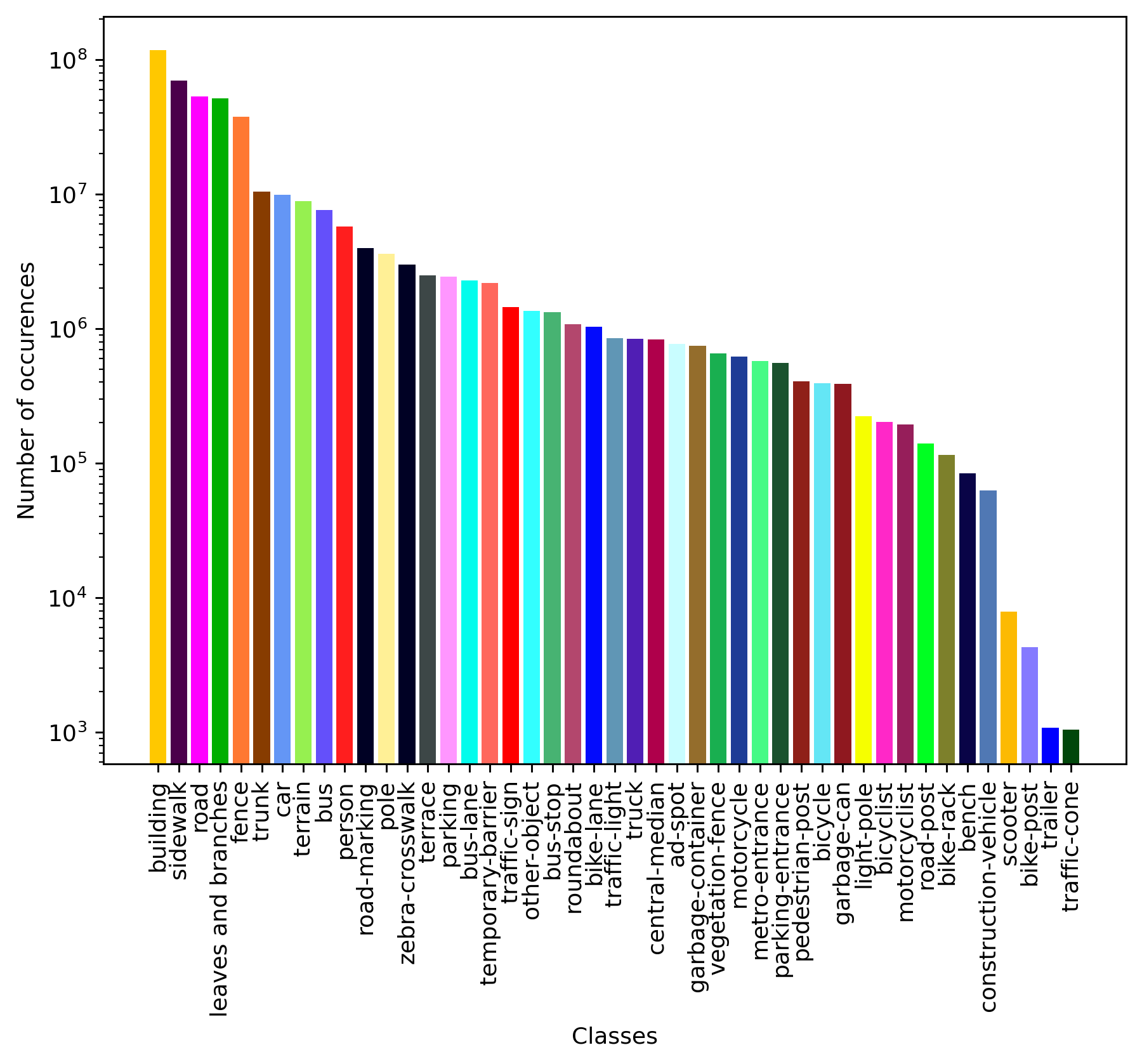}
    \caption{Distribution of labels in ParisLuco3D.}
    \label{fig:distrib}
\end{figure}

Due to the nature of the scene, we observe a large quantity of pedestrians and buses. The details of the label distribution are presented in \autoref{fig:distrib}. 

\subsection{Labels for object detection}

As is the case for most autonomous driving datasets, we mainly focused on road agents and cover them all with our 11 labels. These are \textit{car}, \textit{bus}, \textit{truck}, \textit{trailer}, \textit{bicycle}, \textit{bicyclist}, \textit{motorcycle}, \textit{motorcyclist}, \textit{scooter}, \textit{scootercyclist}, and \textit{pedestrian}. 

We took into account the specificities of the common classes between datasets when choosing our labels and the level of granularity to ease the mapping when working in a cross-dataset setting, which we believe to be relevant for the generalization task. We also seek to be as fine-grained as possible for road users, and base our fine-grained LOD classes on the nuScenes dataset, which are easily mappable to the other datasets. 

Whenever possible, we also distinguish between static objects and dynamic ones, such as \textit{bicycle} and \textit{bicyclist}. 
Further, due to the high redundancy between LiDAR scans, we only annotated one out of ten scans with bounding box annotations, while other scans were also used to help guide the annotation process. As depicted in \autoref{tab:obj_det_ds}, we have the highest number of pedestrians per annotated LiDAR scan compared to other datasets, which are notoriously difficult to detect at long ranges.

\begin{table}[h]
     \centering
    \scriptsize
    \begin{tabular}{c|c|c|c|c|c}
         \multirow{2}{*}{Name} & \multirow{2}{*}{Classes} & \multirow{2}{*}{3D Boxes} &  \multicolumn{3}{c}{\tabcell{Instances per scan}}\\ \cline{4-6}
         & & & car & bicycle & pedestrian \\
         \hline
         \hline
         KITTI \cite{Geiger2012CVPR}& 3 & 18k & 3.8 & 0.2 & 0.6\\
         \hline
         nuScenes\cite{nuscenes} & 23/10$^*$ & 149k & 11& 0.3& 5.7\\
         \hline
         Waymo \cite{waymo}& 3& 1.8M & 29.8 & 0.3 & 13.5\\
         \hline
         ONCE \cite{once_ds}& 5& 90k& 19.8& 6.3& 2.9\\
         \hline
         \hline
         ParisLuco3D & 11/7/5$^{**}$& 64k& 14.7& 6.4& 56.3\\
         
    \end{tabular}
    \caption{Comparison between 3D object detection datasets. All values are computed for the validation splits of the datasets. $^*$The 23 original classes of nuScenes are grouped during evaluation into 10 due to a strong similarity in a few fine classes---for example, \textit{standing pedestrians} and \textit{sitting pedestrians}. $^{**}$While we annotate 11 different classes, we evaluate 7 of these classes for models trained on nuScenes and 5 for those trained on ONCE, using the intersection of classes between these datasets and ours.}
    \label{tab:obj_det_ds}
\end{table}
Due to the natural occlusion phenomenon that occurs during traversal of the environment, certain annotated objects may not contain any points. However, if during the annotation process we became aware of the presence of an object, for example by aggregating sequential point clouds, we maintained its annotation in order to facilitate both annotation and the tracking task.

While we annotated bounding boxes to their maximum range, certain filters are applied while evaluating them in order to ensure a fair evaluation process. First, only boxes whose center was within 50~\text{m} of the sensor were retained.
Second, since many boxes were kept even if they did not contain any points, we did not expect any model to predict them. In practice, we filter out boxes that contain five points or less for object detection evaluation.

For a given 3D scan containing $N$ annotated objects, bounding box annotations  are represented in the following manner:
$$
\{b_i = (cx_i, cy_i, cz_i, w_i, l_i, h_i, \theta_i)\}_{i \in [1, N] }
$$
where $cx_i$, $cy_i$, and $cz_i$ represent the center of the bounding box annotation along the $x$, $y$, and $z$ coordinates;  $w_i$, $l_i$, and $h_i$ represent the width, length, and height; and $\theta_i$ represents the heading of the bounding box.

\subsection{Labels for tracking}

Every element within the object detection annotations was tracked. This implies that there are also 11 classes for tracking, and one of out 10 scans were annotated. 

While detection computes scores for boxes populated with more than five points, objects that drop below five points due to occlusion are still tracked.
\section{Overview of LiDAR Domain Generalization on ParisLuco3D}

To illustrate the importance of providing a new dataset that specifically targets domain generalization, we benchmarked current state-of-the-art methods of semantic segmentation and object detection, we used ParisLuco3D as the target dataset to avoid the need to compromise on the label set at the time of evaluation.

\subsection{LiDAR semantic segmentation}

\subsubsection{Related work}

LSS builds upon the typical 2D convolutionnal neural networks. Earlier approaches projected the 3D point cloud into a 2D representation, such as a range projection \cite{cenet} or a bird's-eye-view (BEV) projection \cite{polarnet}. While these methods were extremely fast, their performance was not satisfactory

In parallel, point-based architectures were developed by taking the time to redefine the convolution \cite{kpconv}.

The highest performing methods are the sparse voxel-based \cite{mink} models, which restructure the point cloud into a regular 3D grid and apply 3D convolution to it \cite{cylinder3d}.

Recently, a few methods have begun to exploit the acquisition pattern of typical LiDAR sensors to achieve a high inference speed with a limited decrease in performance~\cite{helix4D}.

\subsubsection{Experiments}

To compute domain generalization performance, two datasets are required. One is used as the training set, called the source set, while the other is used as the test set, called the target set. The target set is supposed to have never been seen at any point in the training process. As source sets, we utilized two datasets considered to be standard: SemanticKITTI and nuScenes. They are considered standard due to their size and seniority.
These two source datasets evaluate different domain gaps to the target dataset ParisLuco3D:
\begin{itemize}
\item SemanticKITTI: LiDAR sensor (VelodyneHDL64 $\rightarrow$ VelodyneHDL32 with different positioning in height), environment (suburban $\rightarrow$ urban), country (Germany $\rightarrow$ France)
\item nuScenes: LiDAR sensor (exactly the same LiDAR sensor but different positioning in height), environment (urban $\rightarrow$ urban), country (US+Singapore $\rightarrow$ France).
\end{itemize}

We computed the generalization results for seven different neural architectures from SemanticKITTI to ParisLuco3D and from nuScenes to ParisLuco3D (\autoref{tab:SSDGPL3D}). These different models were selected due to their representativeness of the various input types (projection-based, point-based and voxel-based), the availability of the source code, and the quality of their source-to-source results. All methods were trained with the respective original label sets of SemanticKITTI and nuScenes, and evaluated on that basis.

To date, very few domain generalization methods for LiDAR point clouds have been released. The most notable one \cite{Kim_2023_CVPR} has not made its code available and, thus, we did not benchmark it. Nonetheless, we followed their protocol to create a generalization baseline by using IBN-Net \cite{pan2018two} and RayDrop \cite{theodose:hal-03485613} in addition to SRUNet. IBN-Net relies on modifying the typical convolution block by incorporating instance normalization to reduce the statistical overfitting to the dataset. RayDrop randomly discards the acquisition ring to simulate lower resolution scans in order to increase the variance of the point distribution within a scan.

The metric used for the computation of the semantic segmentation performance is the Intersection over Union (IoU) for each class and its mean (mIoU). 
It is measured at a distance of up to 50 meters away from the sensor, which is similar to SemanticKITTI evaluation.

\begin{table}[h]
    \scriptsize
    \centering
    \setlength\tabcolsep{2pt}
    \begin{tabular}{c|c|c|c|c|c}
    
    &&\multicolumn{2}{c|}{$SK\rightarrow PL$}& \multicolumn{2}{c}{$NS\rightarrow PL$*}\\ \hline
        Model & Input type & mIoU$_{SK}$ & mIoU$_{PL}$&  mIoU$_{NS}$& mIoU$_{PL}$\\\hline\hline
        CENet \cite{cenet}& Range &58.8 &21.6 &69.1&  31.6  \\ \hline
        PolarSeg \cite{polarnet}& BEV & 61.8&9.6 & 71.4 & 11.3 \\ \hline
        KPConv \cite{kpconv}& Point & 61.8 & 20.3& 64.2 &  22.9\\ \hline
        SRUNet \cite{mink}& Voxel & 63.2 &\textbf{30.7} &69.3 & 37.4 \\ \hline 
        SPVCNN \cite{spvnas}& \tabcell{Point+Voxel} & 63.4 & 28.9& 66.8&  \textbf{38.7}\\ \hline
        Helix4D \cite{helix4D}& 4D Point & 63.9& 18.3& 69.3 &  19.2  \\ \hline
        Cylinder3D \cite{cylinder3d}& \tabcell{Cyl. voxel} & \textbf{64.9} &  23.0& \textbf{74.8} & 25.5\\ \hline \hline
       \tabcell{SRUNet \cite{mink}\\+RayDrop \cite{theodose:hal-03485613}}&Voxel& 61.7 & \textbf{35.8} &66.4 & 31.6  \\ \hline
       \tabcell{SRUNet \cite{mink}\\+IBN-Net \cite{pan2018two}}&Voxel & \textbf{64.9} & 28.0& \textbf{67.3} & \textbf{39.7}\\ 

    \end{tabular}
    \caption{Generalization baseline when trained on SemanticKITTI~(SK) or nuScenes (NS), and evaluated on ParisLuco3D (PL). *When nuScenes is the source, the intensity channel of the LiDAR sensor is used.}
    \label{tab:SSDGPL3D}
\end{table}
\vspace{-5mm}
\subsubsection{Results and analysis}

\autoref{tab:SSDGPL3D} first shows that good performance of a neural architecture in source-to-source does not imply good performance in target at all. In addition, generalization methods (Raydrop, IBN-Net) do not systematically allow a performance gain on the target.
\autoref{tab:SSDGPL3D} indicates that the best performing generalization neural architecture is the voxel-based one---namely SRUNet~\cite{mink}. This is why we selected this architecture to test the IBN-Net and RayDrop generalization methods. But overall, the performance of all architectures on ParisLuco3D is disappointing compared to their performance on the training set.


For SemanticKITTI to ParisLuco3D, all architectures were retrained without the intensity channel. As the sensors used in the two datasets varied, the intensity decreased the generalization performance. In \autoref{tab:reflectivity}, we illustrate this decrease in performance in several cases.

For nuScenes to ParisLuco3D, the results were computed with the intensity channel, as the same LiDAR sensor is used in both datasets. This is also illustrated in \autoref{tab:reflectivity}, where we can see that the results are better with than without intensity. The low mIoU (39.7 mIoU$_{PL}$ for the best model) comes from the \textit{trailer} and \textit{traffic-cone} categories for which there are very few examples in ParisLuco3D. It is important to keep these minority classes because they allow models to be evaluated on the rare edge cases.

\begin{table}[h]
    \scriptsize
    \centering
    \begin{tabular}{c|c|c|c|c}
    
    & \multicolumn{2}{c|}{$SK\rightarrow PL$}& \multicolumn{2}{c}{$NS\rightarrow PL$}\\ \hline
        Model & mIoU$_{SK}$ & mIoU$_{PL}$&  mIoU$_{NS}$& mIoU$_{PL}$\\\hline\hline
        \tabcell{SRUNet \cite{mink}\\
        with intensity}& 66.6 & 27.9 & 69.3 & 37.4 \\ \hline 
        \tabcell{SRUNet \cite{mink}\\
        without intensity} & 63.2 &30.7 & 66.3 & 32.3\\ \hline \hline
        \tabcell{Cylinder3D \cite{cylinder3d}\\
        with intensity}& 70.4 &2.7 & 74.8 & 25.5\\ \hline 
        \tabcell{Cylinder3D \cite{cylinder3d}\\
        without intensity} & 64.9 & 23.0& 70.2 &17.1 \\
    \end{tabular}
    \caption{Impact of the intensity channel on generalization performance for semantic segmentation task.}
    \label{tab:reflectivity}
\end{table}

With regards to domain generalization methods, we can observe that IBN-Net is useful for nuScenes towards ParisLuco3D and detrimental for SemanticKITTI. It stems for the increased similarity in instance statistics between nuScenes and ParisLuco3D due to the intensity channel. 
Conversely, RayDrop is useful in the SemanticKITTI towards ParisLuco3D case. Using RayDrop helps the model learn to segment a larger variety of lower sensor resolution. Due to the inability of methods to faithfully increase the resolution, it is detrimental when the original resolution is lower than the target resolution, which is the case for nuScenes (LiDAR rotating at 20Hz) towards ParisLuco3D (same LiDAR rotating at 10Hz).

\subsection{LiDAR object detection}
\subsubsection{Related work}
LOD predicts bounding boxes by applying local shape feature extractors and using convolutional filters similarly to 2D object detection.
Works such as SECOND \cite{Yan_Mao_Li_2018} and PointPillars \cite{Lang_2019_CVPR} sought to increase the speed of using these convolutions to enable real-time applications.

However, point-based methods also exist. For example, PointRCNN~\cite{Shi_2019_CVPR} obtains point-wise multi-range features to optimize predictions.

The best performing methods typically use two-steps processes of initial proposals and refinement using features interpolated from keypoints~\cite{Shi_2020_CVPR, Yin_2021_CVPR}.
Recently, methods have been exploring predictions based on object centers~\cite{Yin_2021_CVPR}\change{, sparse features {\cite{chen2023voxenext}}} as well as transformer-based architectures~\cite{Wang_2023_CVPR}.

\subsubsection{Experiments}
We established a benchmark for testing models on our ParisLuco3D dataset. As we aimed to test the domain generalization performance of detection models, we trained them on two standard source datasets: ONCE and nuScenes. Then, we evaluated them on ParisLuco3D, using the latter as a test set. Therefore, ParisLuco3D scans and annotations were not utilized during the training process. We chose ONCE due to its size and its differences in LiDAR sensors, a 40-beam sensor. The source dataset ONCE evaluates different domain gaps to the target dataset ParisLuco3D:
\begin{itemize}
\item ONCE: LiDAR sensor (40-beam LiDAR $\rightarrow$ VelodyneHDL32 with different positioning in height), environment (sub.+urban $\rightarrow$ urban), country (China $\rightarrow$ France)
\end{itemize}

We evaluated the generalization ability of five different 3D object detection models across the various choices of source datasets: SECOND \cite{Yan_Mao_Li_2018}, PointRCNN \cite{Shi_2019_CVPR}, PointPillars \cite{Lang_2019_CVPR}, PV-RCNN \cite{Shi_2020_CVPR}, \changedel{and }CenterPoint \cite{Yin_2021_CVPR}\change{, VoxelNeXt {\cite{chen2023voxenext}}, and DSVT {\cite{Wang_2023_CVPR}}}. These models were selected due to their availability and their performance across different 3D object detection benchmarks. \change{Notably, VoxelNeXt and DSVT are the current state-of-the-art 3D object detection models for the Argoverse2 {\cite{wilson2021argoverse}} dataset and nuScenes {\cite{nuscenes}} dataset respectively.}


We also implement the same methods for domain generalization as previously used for semantic segmentation---that is instance normalization IBN-Net \cite{pan2018two} and ray-dropping data augmentations inspired by \andal{Theodose} \cite{theodose:hal-03485613}.


The predictions were evaluated by considering those with low overlap with ground-truth bounding boxes as false positives. We used IoU thresholds specific to each class for this---specifically 0.7 for four-wheel vehicles, 0.5 for two-wheel vehicles, and 0.3 for pedestrians. 
Furthermore, we use the average precision (AP) metric, which is commonly used for the 3D object detection task, and its mean across all classes (mAP). The average precision for a given class is defined as the precision integrated over a fixed range of recall values. We selected 50 recall values ranging between 0.02 and 1, following the protocol of ONCE.

\subsubsection{Results \& analysis}

\begin{table}[h]
    \scriptsize
    \centering
    \scalebox{1}{%
    \begin{tabular}{c|c|c|c|c}
    & \multicolumn{2}{c|}{$ON\rightarrow PL$} & \multicolumn{2}{c}{$NS\rightarrow PL$} \\ \hline
        Model & mAP$_{ON}$ & mAP$_{PL}$ &  mAP$_{NS}$& mAP$_{PL}$\\\hline\hline
        PointRCNN \cite{Shi_2019_CVPR} & 28.6 & 8.3& 18.4 & 10.6\\ \hline
        PointPillars \cite{Lang_2019_CVPR} & 45.5 & 12.4  & 35.3 & 10.9\\ \hline
        PV-RCNN \cite{Shi_2020_CVPR} & 52.4 & 10.6 & 29.6 &\textbf{17.3}\\ \hline
        SECOND \cite{Yan_Mao_Li_2018} & 54.0 & 9.9 & 38.6 & 12.9\\ \hline 
        CenterPoint \cite{Yin_2021_CVPR} & 59.5 &8.5  & 35.0 & 13.4\\  \hline 
        \change{VoxelNeXt {\cite{chen2023voxenext}}}&  \change{32.2}& \change{5.2} &  \change{42.2}& \change{11.1}\\ \hline \hline
        \change{DSVT {\cite{Wang_2023_CVPR}}}&  \change{\textbf{64.4}}&  \change{\textbf{16.9}}&  \change{\textbf{44.2}}& \change{9.1}\\ \hline \hline
        \tabcell{PointPillars \cite{Lang_2019_CVPR}\\+RayDrop \cite{theodose:hal-03485613}} &  39.4&  \textbf{13.3} & 29.8 & 9.5\\ \hline 
        \tabcell{PointPillars \cite{Lang_2019_CVPR}\\+IBN-Net \cite{pan2018two}} &  \textbf{45.0} &  11.1& \textbf{35.9} & \textbf{10.3}\\  
    \end{tabular}
    }
    \caption{Generalization baseline for LOD when trained on ONCE (ON) and nuScenes (NS) and evaluated on ParisLuco3D (PL). The intensity channel is not used.}
    \label{tab:once_res}
\end{table}

We present our benchmark results in \autoref{tab:once_res} in terms of mAP.
Our models were trained without the intensity channel, as we emphasize in \autoref{tab:reflectivity_objdet} how intensity hinders the generalization ability of models. In \autoref{tab:once_res}, despite its relatively low source mAP, the PointPillars models tends to generalize better on ParisLuco3D when trained on ONCE. We believe that this is because it contains a lower number of parameters, and so it tends to be less susceptible to overfitting. On the other hand, PV-RCNN works best when training using nuScenes, as the model is natively capable of capturing more fine-grained details, which is necessary when working on the low resolution of nuScenes. \change{Finally, we find that though DSVT achieves high source accuracy on nuScenes when using close to optimal hyperparameters, it suffers the most from generalization accuracy degradation on our dataset.}


\autoref{tab:once_res} highlights the fact both RayDrop and IBN-Net are insufficient generalization baselines and their impact may depend on the source dataset used. For example, using the RayDrop augmentation for a PointPillars model trained on ONCE when the source LiDAR density is higher than that of ParisLuco3D tends to help the generalization ability of this model ($+0.9$ mAP). However, the opposite occurs when we use RayDrop with nuScenes \changedel{($-1.2$ mAP)}\change{($-1.4$ mAP)}. RayDrop increases the generalization for most classes for our PointPillars model trained on ONCE, making it the model with the \change{second} highest mAP on ParisLuco3D despite having one of the lowest mAP performances on its own source datasets. We believe this highlights the fact that aligning the LiDAR resolutions while taking advantage of large datasets can be a strong generalization method. However, as shown using nuScenes as a training set, it is not a one size fits all method and is dependant on the difference in LiDAR sensor resolutions between a source and a target dataset, which is normally unknown in the conditions of the domain generalization task.

\begin{table}[h]
    \scriptsize
    \centering
    \begin{tabular}{c|c|c|c|c}
    
    & \multicolumn{2}{c|}{$ON\rightarrow PL$}& \multicolumn{2}{c}{$NS\rightarrow PL$}\\ \hline
        Model & mAP$_{ON}$ & mAP$_{PL}$&  mAP$_{NS}$& mAP$_{PL}$\\\hline\hline
        \tabcell{PointPillars \cite{Lang_2019_CVPR}\\
        with intensity} &  45.0& 9.2&  40.9& 2.5\\  \hline 
        \tabcell{PointPillars \cite{Lang_2019_CVPR}\\
        without intensity} & 45.5 &  12.4&  \changedel{37.7}\change{35.3}& \changedel{12.7}\change{10.9}\\  \hline \hline 
        \tabcell{CenterPoint \cite{Yin_2021_CVPR}\\
        with intensity} &  62.5& 6.5& 41.1 & 10.5\\ \hline 
        \tabcell{CenterPoint \cite{Yin_2021_CVPR}\\
        without intensity} &  59.5&  8.5&  \changedel{38.7}\change{35.0}& \changedel{15.6}\change{13.4}\\ \hline \hline
        \tabcell{\change{DSVT} \cite{Wang_2023_CVPR}\\
        \change{with intensity}} & \change{65.0} & \change{13.7} & \change{47.6}& \change{3.5}\\ \hline 
        \tabcell{\change{DSVT} \cite{Wang_2023_CVPR}\\
        \change{without intensity}} & \change{64.4} & \change{16.9} &  \change{44.2}& \change{9.1}\\
    \end{tabular}
    \caption{Impact of the intensity channel on generalization performance for the detection task.}
    \label{tab:reflectivity_objdet}
\end{table}

In \autoref{tab:reflectivity_objdet}, we trained \changedel{SECOND, CenterPoint, and PointPillars} \change{PointPillars, CenterPoint, and DSVT} on the ONCE and nuScenes datasets both with and without intensity. It is evident that unlike in semantic segmentation, using intensity for object detection when training on nuScenes actually hinders the final performance by a large amount, ranging from a \changedel{+0.9} \change{-2.0} to -10.2 mAP difference on ParisLuco3D compared to models trained without intensity (yet nuScenes and ParisLuco3D have the same LiDAR sensor). We believe this is due to a difference in the intensity distributions of objects, which has been shown to be a major factor of model overfitting for LOD \cite{Schinagl_2022_CVPR}.

Overall, we found that both LSS and LOD models suffer from large accuracy drops when transferred to our ParisLuco3D dataset, thereby highlighting the need for novel neural architectures and specific 3D generalization methods. The intensity channel of LiDAR is an input which is a source of overfitting in source-to-source and requires further research to be better exploited in domain generalization (even with the same LiDAR sensor as target).

\section{Online benchmark}

To ensure a fair measurement of the domain generalization performance on this dataset, only the raw dataset is released and not the ground-truth labels. For qualitative comparisons of methods, five scans have been released alongside their annotations.

We then establish online benchmarks to test the domain generalization for three LiDAR perception tasks: semantic segmentation, object detection, and object tracking.


All details for online benchmarks are available on the project website: \url{https://npm3d.fr/parisluco3d}




We encourage authors to avoid typical benchmark optimization techniques in order to fairly judge generalization performance. Following \cite{torralba2011unbiased}, the number of submissions is limited.

\section{Conclusion}

In this paper, we proposed a new dataset for LiDAR perception. This dataset stands out from other already released datasets due to the fineness of the annotations, and its goal of domain generalization evaluation.

We also benchmarked different neural architectures and different generalization methods for LiDAR semantic segmentation and LiDAR object detection, thereby demonstrating that the most recent and efficient architectures are not the most robust and that generalization methods have mixed results for now.

We hope that this dataset will enable visibility in our community regarding the emergence of new 3D perception methods that generalize well, thereby allowing 3D perception tasks to be used in real-world conditions.
\section*{Acknowledgments}
We would like to thank Hassan Bouchiba, who mainly contributed to the acquisition of this dataset back in 2015.

{\small
\bibliographystyle{IEEEtran}
\bibliography{IEEEabrv,bib/dataset,bib/method,bib/LSS, bib/misc}
}
\clearpage
\appendix

\subsection{Experiment details}

\subsubsection{Parameter set for LSS}

Studied models for LSS, SalsaNext, PolarSeg, KPConv, Helix4D, Cylinder3D, SRUNet, and SPVCNN, were taken from their respective official github and trained with the standard hyperparameters shared by the authors.

For IBN-Net, we apply instance normalization at the same position as \cite{Kim_2023_CVPR}.

For PolarMix, we use the official repository, and apply scene mixing with a probability of 0.5 and object mixing with a probability of 0.5.

The LSS annotation is guaranteed between 5 m and 50 m away from the sensor, and as such, the mIoU is computed within this range.

\subsubsection{Parameter set for LOD}
Studied models for LOD, SECOND, PointRCNN, PointPillars, PV-RCNN, and CenterPoint, were taken from the framework OpenPCDet\footnote{https://github.com/open-mmlab/OpenPCDet} and trained with the standard hyperparameters for 30 epochs.

For IBN-Net, we apply an instance normalization layer to both the 3D backbone of PV-RCNN as well as its compressed 2D Bird's Eye View (BEV) map of features. For the 3D backbone, instance normalization is applied after the first, second to last and last convolutional blocks, before the ReLU activation layer. For the 2D BEV map, instance normalization is also applied in that same order.

In the case of RayDrop, we drop between 25 and 60 percent of LiDAR beams to emulate lower-resolution sensors, similarly to \andal{Theodose}~\cite{theodose:hal-03485613}.

All predictions at transfer time are filtered according to a maximum range of 50~\text{m}, predictions at the same location are also filtered using a NMS algorithm with a 0.1 IoU threshold. As previously mentioned, we also filter ground truth bounding box during evaluation to only keep boxes with strictly more than 5 points contained within them.

\subsection{Labels details}

\subsubsection{LSS labels}
\begin{figure}
    \centering
    \captionsetup[subfigure]{justification=centering}
        \subfloat[Bus on a bus lane]{
            \includegraphics[width=0.3\linewidth]{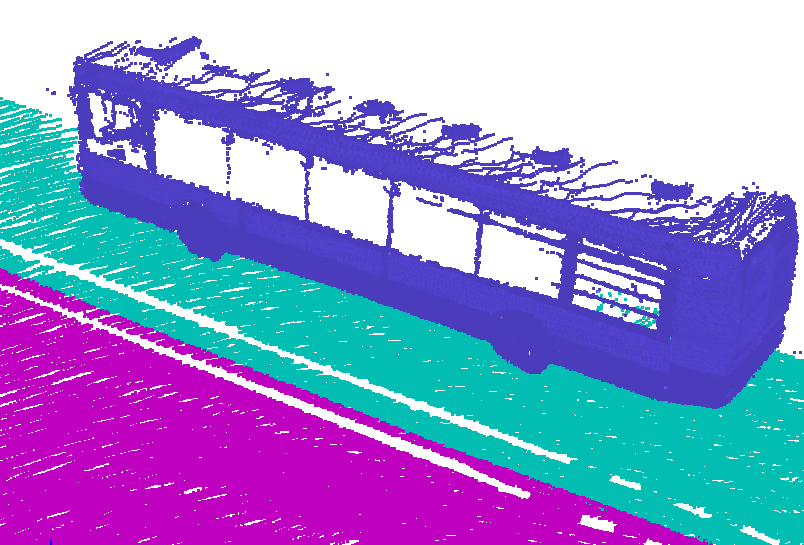}
        }
        \subfloat[Bus stop]{
            \includegraphics[width=.3\linewidth]{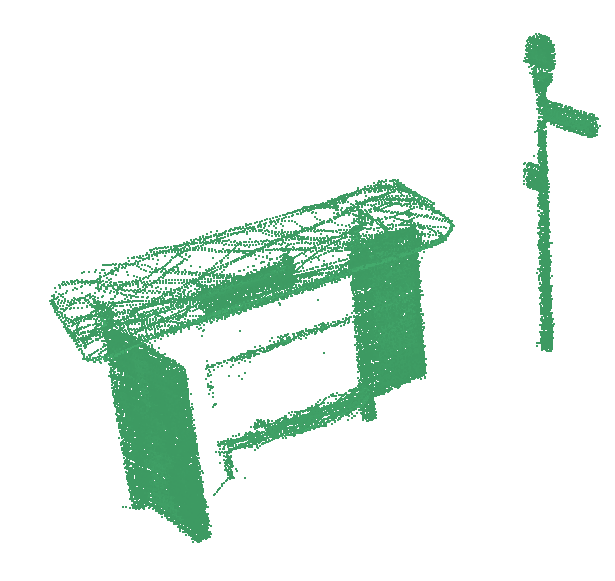}
        }
        \subfloat[Road post on central median]{
            \includegraphics[width=.2\linewidth]{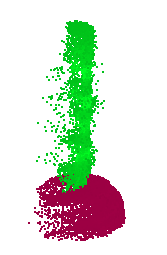}
        }\\
        \subfloat[Road leading to a parking entrance. Road markings and Zebra crosswalk]{
            \includegraphics[width=0.4\linewidth]{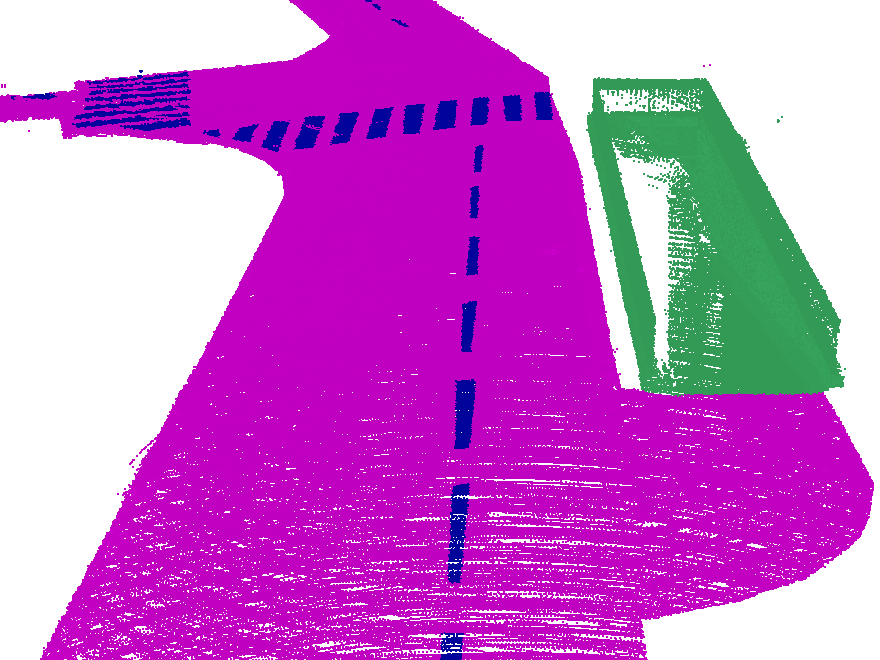}
        }
        \hspace{1em}
        \subfloat[Roundabout with an other object in the middle]{
            \includegraphics[width=.3\linewidth]{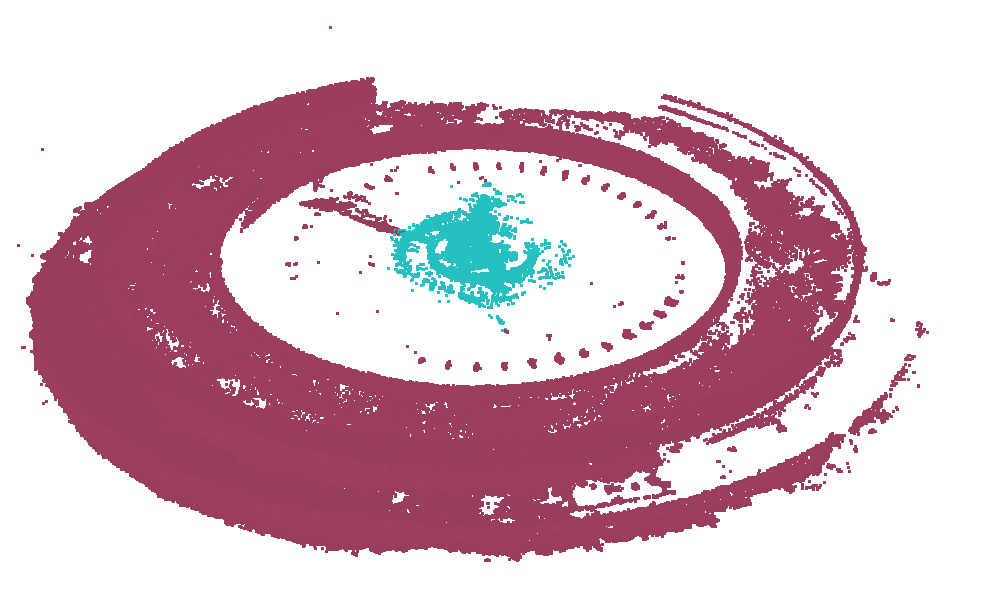}
        }\\
        \subfloat[Bench]{
            \includegraphics[width=0.3\linewidth]{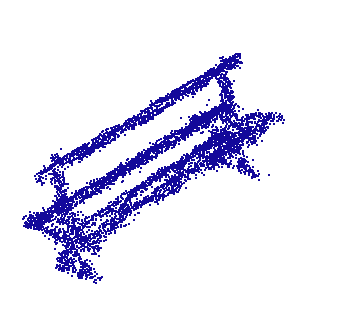}
        }
        \subfloat[Car on parking]{
            \includegraphics[width=.3\linewidth]{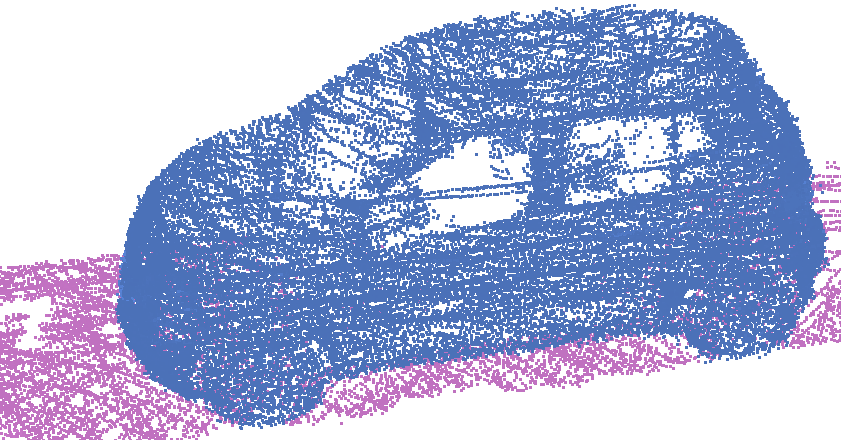}
        }
        \subfloat[Construction vehicle]{
            \includegraphics[width=.3\linewidth]{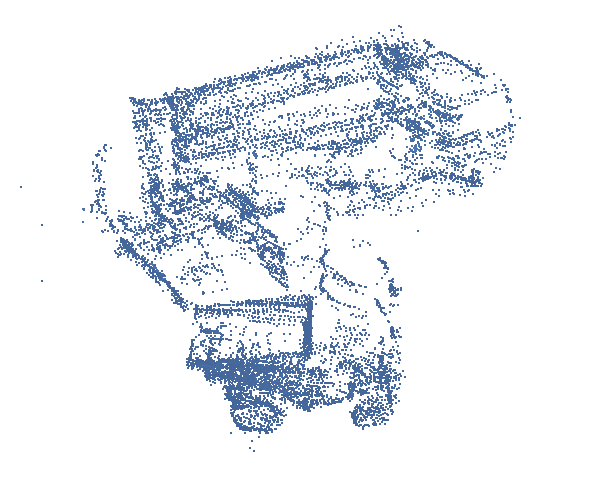}
        }\\

        \subfloat[Fence]{
            \includegraphics[width=0.4\linewidth]{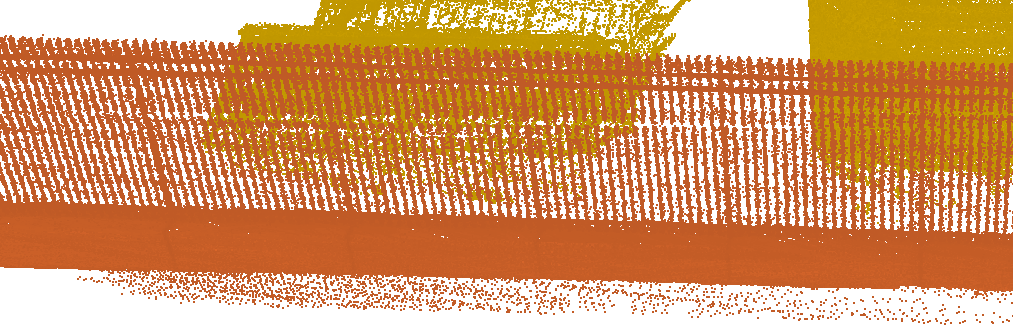}
        }
        \subfloat[Garbage can]{
            \includegraphics[width=.2\linewidth]{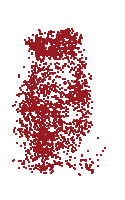}
        }
        \subfloat[Garbage container]{
            \includegraphics[width=.3\linewidth]{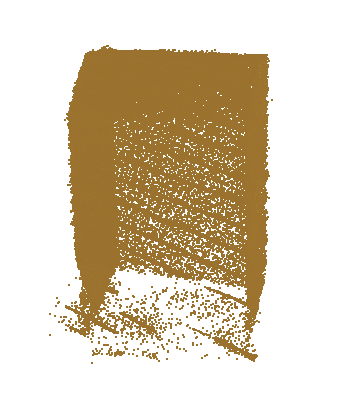}
        }\\

        \subfloat[Metro entrance]{
            \includegraphics[width=0.3\linewidth]{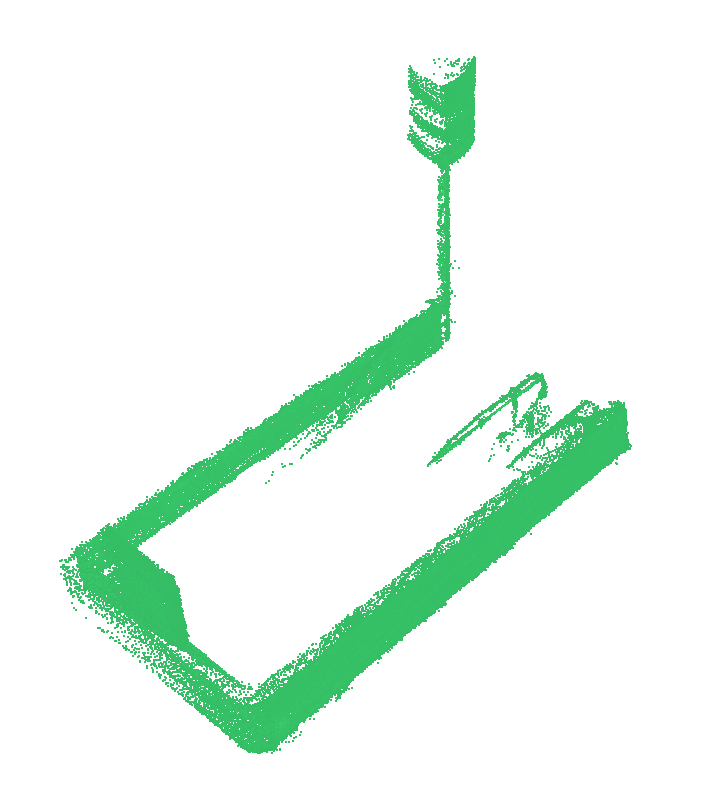}
        }
        \subfloat[Pedestrian on sidewalk]{
            \includegraphics[width=.2\linewidth]{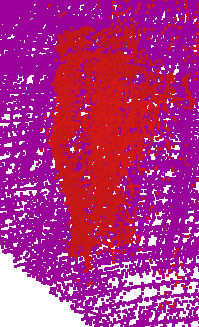}
        }
        \subfloat[Building]{
            \includegraphics[width=.4\linewidth]{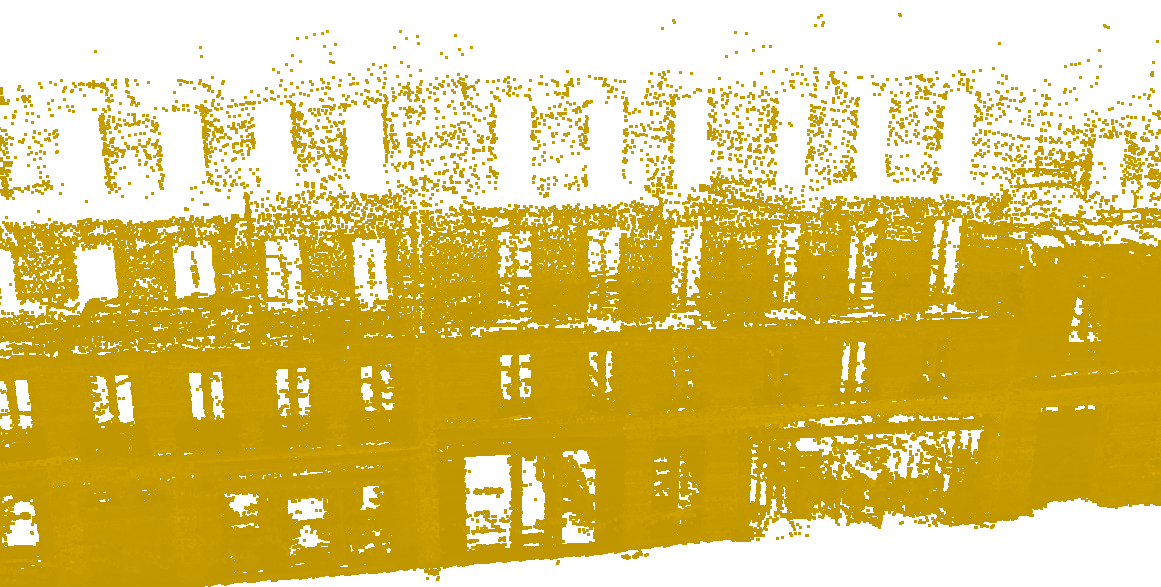}
        }
    \caption{Labels for LSS.}
    \label{fig:lss_labl_1}
\end{figure}

\begin{figure}
    \centering
    \captionsetup[subfigure]{justification=centering}
        \subfloat[Trunk topped with vegetation, on terrain]{
            \includegraphics[width=.2\linewidth]{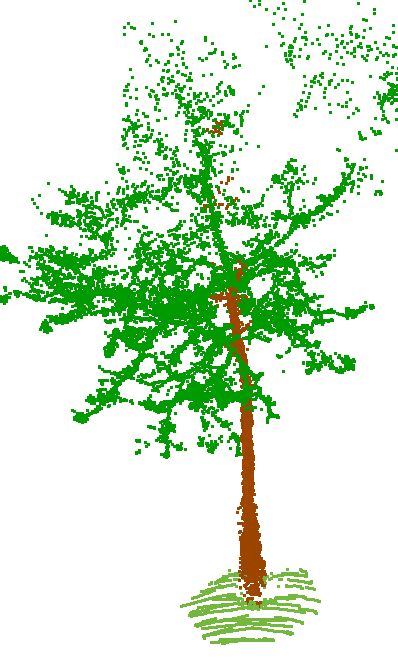}
        }
        \hspace{1em}
        \subfloat[Bicycle and motorcycle with bike rack]{
            \includegraphics[width=0.45\linewidth]{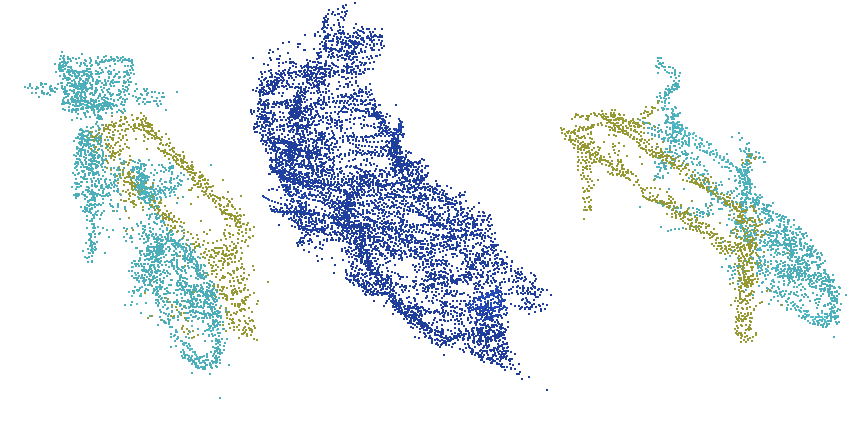}
        }
        \subfloat[Scooter]{
            \includegraphics[width=.2\linewidth]{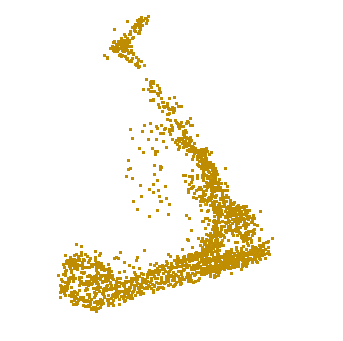}
        }\\
        \subfloat[Sign and light pole on pole]{
            \includegraphics[width=.2\linewidth]{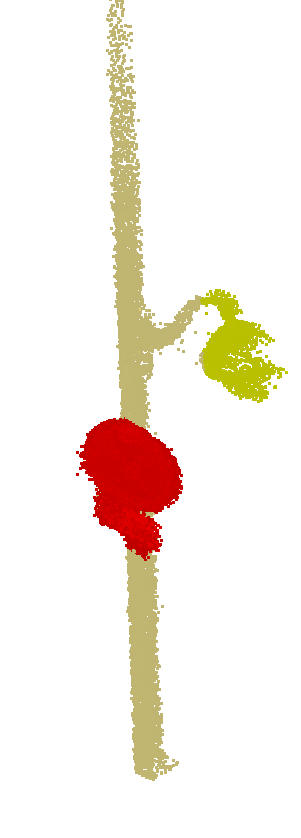}
        }
        \hspace{1em}
        \subfloat[Bike post seperating road from a bike lane]{
            \includegraphics[width=.3\linewidth]{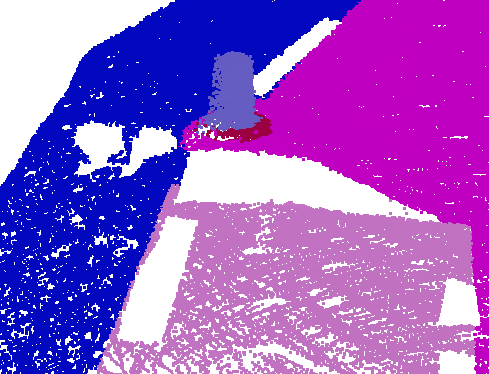}
        }
        \hspace{1em}
        \subfloat[Traffic cone and temporary barrier]{
            \includegraphics[width=.3\linewidth]{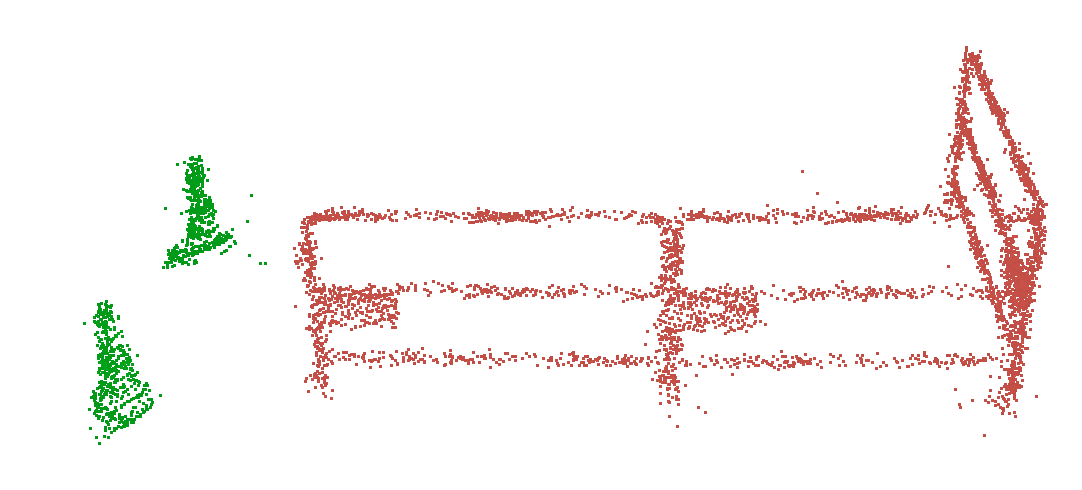}
        }\\
        \subfloat[Terrace]{
            \includegraphics[width=.3\linewidth]{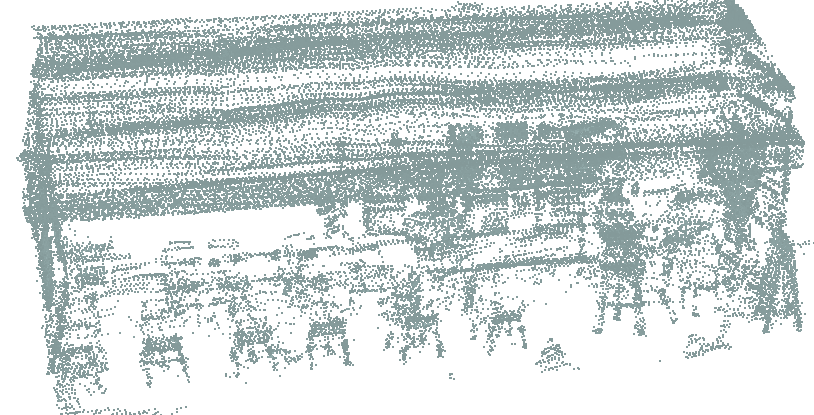}
        }
        \subfloat[Traffic light on a pole]{
            \includegraphics[width=.1\linewidth]{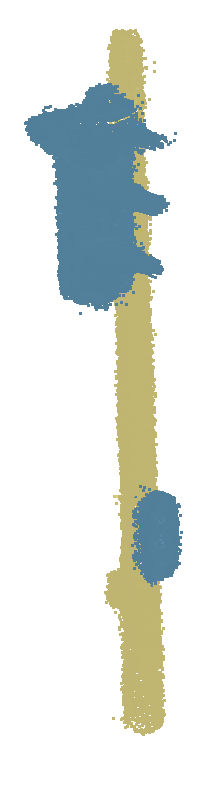}
        }
        \subfloat[Trailer]{
            \includegraphics[width=.5\linewidth]{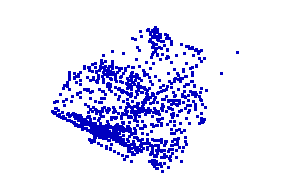}
        }\\
        \subfloat[Truck]{
            \includegraphics[width=.3\linewidth]{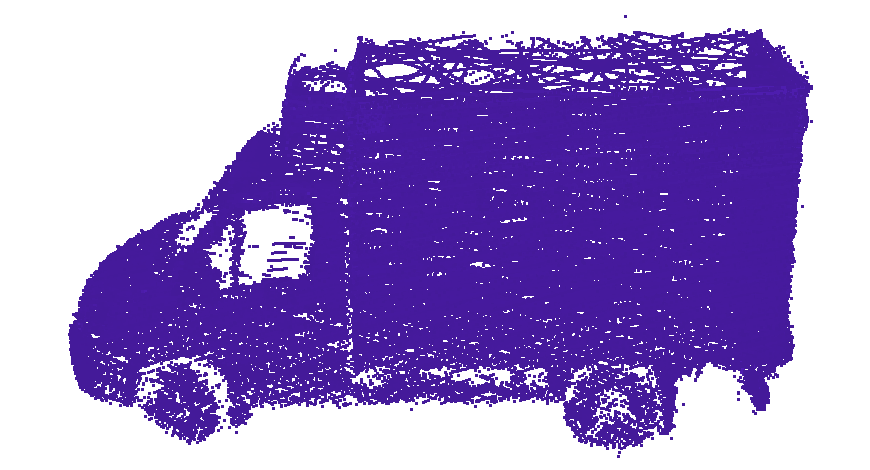}
        }
        \subfloat[Ad Spot]{
            \includegraphics[width=.2\linewidth]{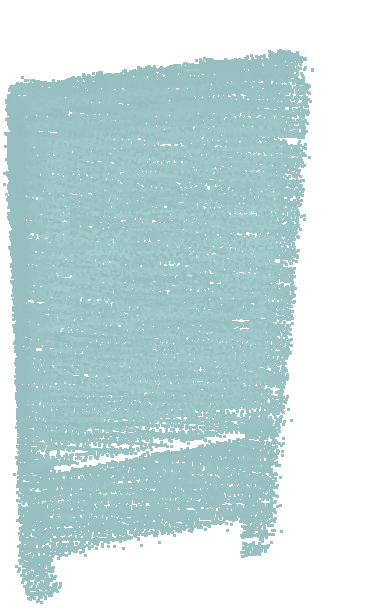}
        }
        \subfloat[Pedestrian post]{
            \includegraphics[width=.2\linewidth]{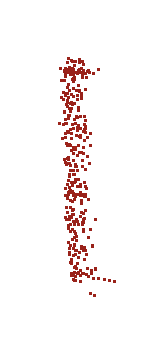}
        }\\
        \subfloat[Trail of a bicyclist]{
            \includegraphics[width=.2\linewidth]{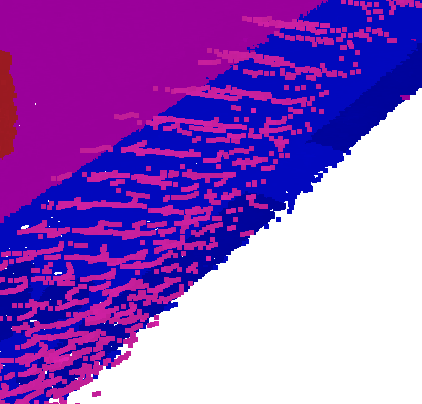}
        }
        \subfloat[trail of a motorcyclist]{
            \includegraphics[width=.4\linewidth]{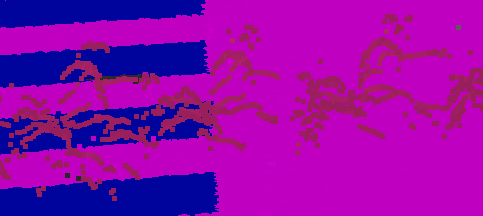}
        }
    \caption{Labels for LSS.}
    \label{fig:lss_labl_2}
\end{figure}

Examples of labels of LiDAR Semantic Segmentation of our dataset ParisLuco3D are shown in \autoref{fig:lss_labl_1} and \autoref{fig:lss_labl_2}.

\subsubsection{LOD labels}
\begin{figure}
    \centering
    \captionsetup[subfigure]{justification=centering}
        \subfloat[car]{
            \includegraphics[width=0.3\linewidth]{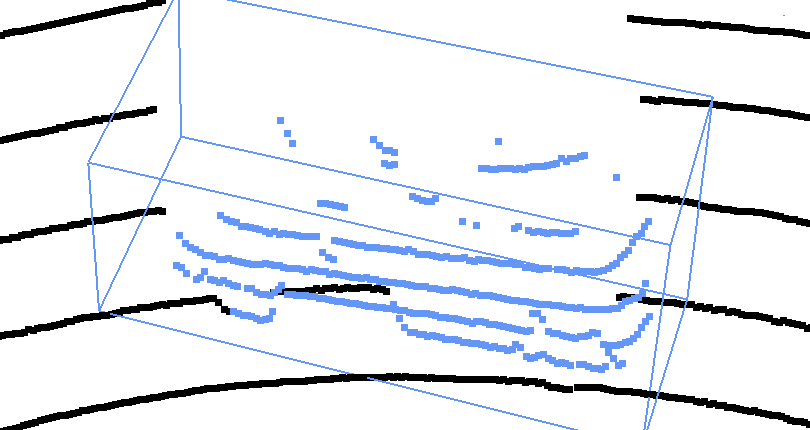}
        }
        \subfloat[motorcycle]{
            \includegraphics[width=.3\linewidth]{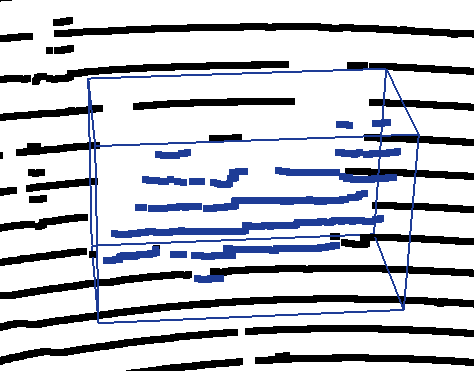}
        }
        \subfloat[bicycle]{
            \includegraphics[width=.3\linewidth]{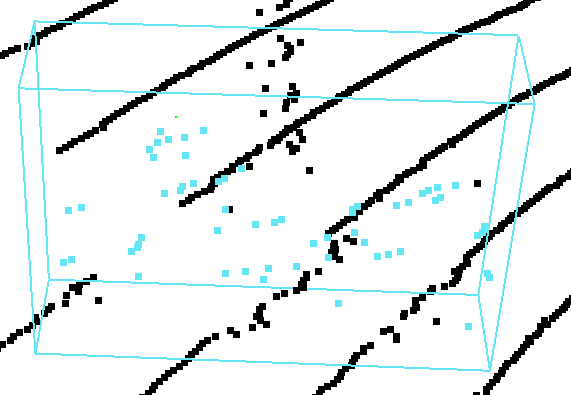}
        }\\
        \subfloat[motorcyclist]{
            \includegraphics[width=.3\linewidth]{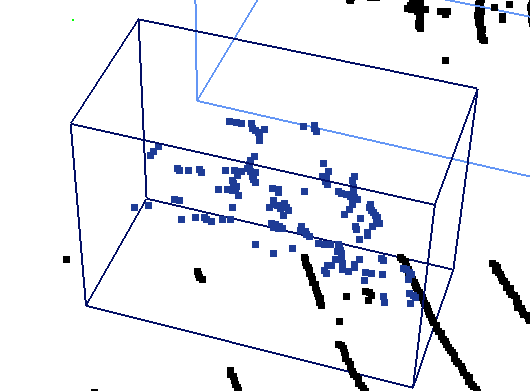}
        }
        \subfloat[bicyclist]{
            \includegraphics[width=.3\linewidth]{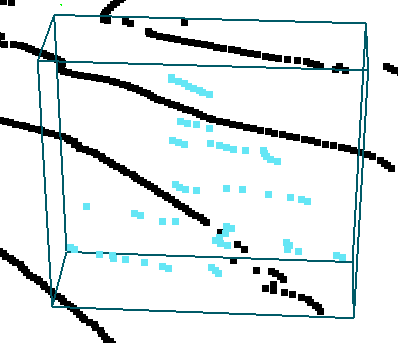}
        }
        \subfloat[truck]{
            \includegraphics[width=.3\linewidth]{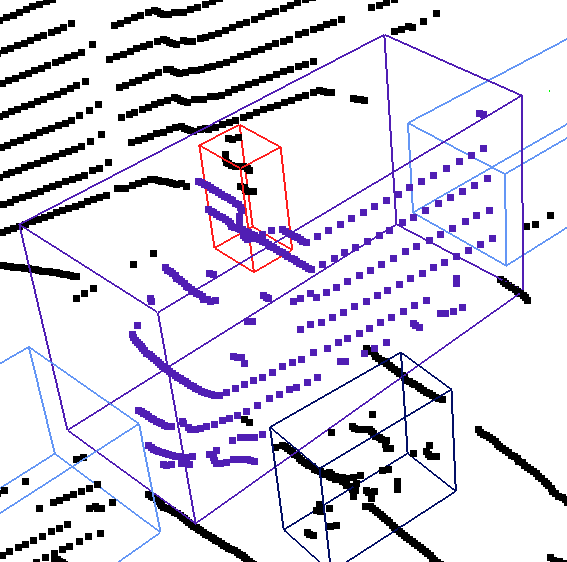}
        }\\
        \subfloat[trailer]{
            \includegraphics[width=.3\linewidth]{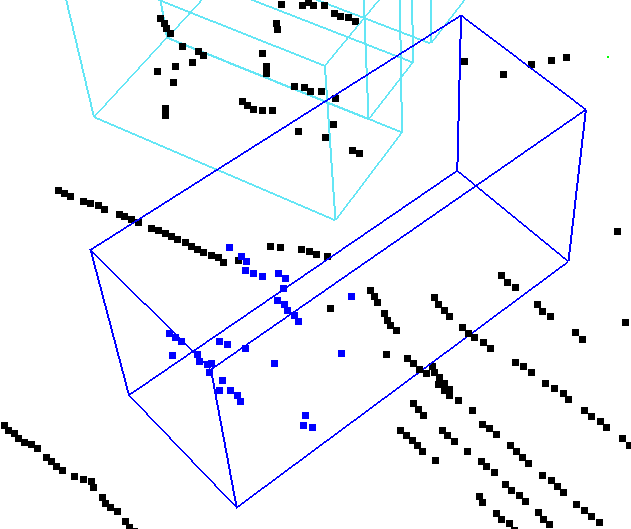}
        }
        \subfloat[bus]{
            \includegraphics[width=.5\linewidth]{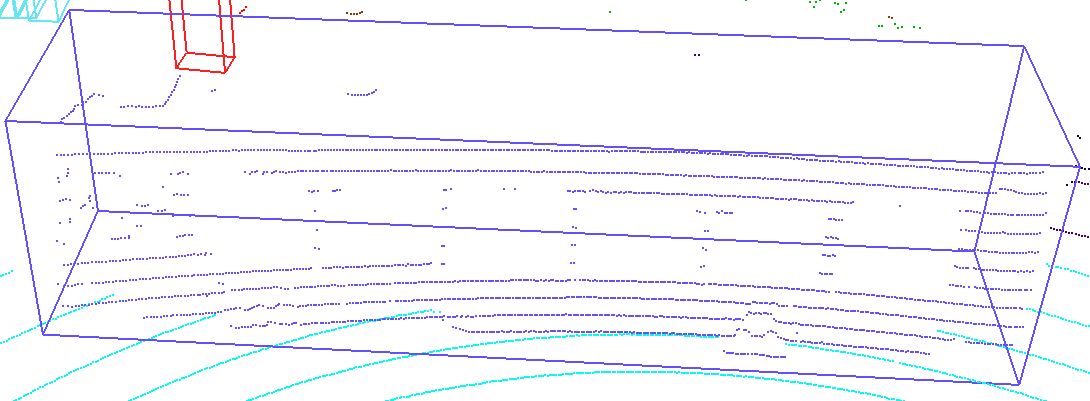}
        }\\
        \subfloat[pedestrian]{
            \includegraphics[width=.2\linewidth]{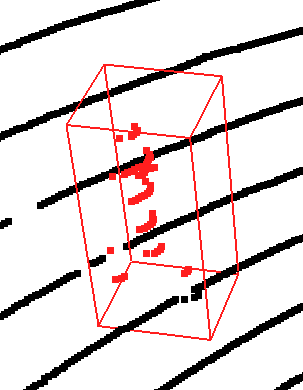}
        }
        \subfloat[scooter]{
            \includegraphics[width=.3\linewidth]{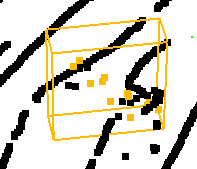}
        }
    \caption{Labels for LOD. While scootercyclist was annotated, none were found in the dataset.}
    \label{fig:lod_labl}
\end{figure}

Examples of labels of LiDAR Object Detection and Tracking of our dataset ParisLuco3D are shown in \autoref{fig:lod_labl}.

\subsection{Remapping details}

\subsubsection{Label mapping for LSS}

\paragraph{SemanticKITTI to ParisLuco3D}
We use the following mapping for the SemanticKITTI dataset to our ParisLuco3D labels, unmentioned labels are discarded:
\begin{itemize}
    \item car $\rightarrow$ car
    \item bicycle $\rightarrow$ bicycle
    \item motorcyle $\rightarrow$ motorcycle, scooter
    \item truck $\rightarrow$ truck
    \item other-vehicle $\rightarrow$ bus, trailer, construction-vehicle
    \item person $\rightarrow$ person
    \item bicyclist $\rightarrow$ bicyclist
    \item motorcyclist $\rightarrow$ motorcyclist
    \item road $\rightarrow$ road, bus-lane, bike-lane, road-marking, zebra-crosswalk
    \item parking $\rightarrow$ parking
    \item sidewalk $\rightarrow$ sidewalk
    \item other-ground $\rightarrow$ roundabout, central-median
    \item building $\rightarrow$ building
    \item fence $\rightarrow$ fence, temporary-barrier
    \item vegetation $\rightarrow$ vegetation
    \item trunk $\rightarrow$ trunk
    \item terrain $\rightarrow$ terrain
    \item pole $\rightarrow$ pole, light-pole
    \item traffic-sign $\rightarrow$ traffic-sign, traffic-light
\end{itemize}

\paragraph{nuScenes to ParisLuco3D}
We use the following mapping for the nuScenes dataset to our ParisLuco3D labels, unmentioned labels are discarded:
\begin{itemize}
    \item barrier $\rightarrow$ temporary-barrier
    \item bicycle $\rightarrow$ bicycle, bicyclist
    \item bus $\rightarrow$ bus
    \item car $\rightarrow$ car
    \item construction-vehicle $\rightarrow$ construction-vehicle
    \item motorcycle $\rightarrow$ motorcycle, scooter, motorcyclist
    \item pedestrian $\rightarrow$ pedestrian
    \item traffic-cone $\rightarrow$ traffic-cone
    \item trailer $\rightarrow$ trailer
    \item truck $\rightarrow$ truck
    \item driveable-surface $\rightarrow$ road, bus-lane, parking, road-marking, zebra-crosswalk
    \item other-flat $\rightarrow$ roundabout, central-median
    \item sidewalk $\rightarrow$ sidewalk, bike-lane
    \item terrain $\rightarrow$ terrain
    \item manmade $\rightarrow$ building, fence, pole, traffic-sign, bus-stop, traffic-light, light-pole
    \item vegetation $\rightarrow$ vegetation, trunk
\end{itemize}

\subsubsection{Label mapping for LOD}

\paragraph{ONCE to ParisLuco3D}
We use the following mapping for the ONCE dataset to our ParisLuco3D labels:
\begin{itemize}
    \item car $\rightarrow$ car
    \item bus $\rightarrow$ bus
    \item truck $\rightarrow$ truck
    \item cyclist $\rightarrow$ bicycle, bicyclist, motorcycle, motorcyclist
    \item pedestrian $\rightarrow$ pedestrian
\end{itemize}
We ignore during the evaluation our classes, such as \textit{trailer}, that are not present in ONCE. \newline

\paragraph{nuScenes to ParisLuco3D}
We use the following mapping for the nuScenes dataset to our ParisLuco3D labels:
\begin{itemize}
    \item car $\rightarrow$ car
    \item bus $\rightarrow$ bus
    \item truck $\rightarrow$ truck
    \item trailer $\rightarrow$ trailer
    \item bicycle $\rightarrow$ bicycle, bicyclist
    \item motorcycle $\rightarrow$ motorcycle, motorcyclist
    \item pedestrian $\rightarrow$ pedestrian
\end{itemize}
Some of the classes from nuScenes, such as \textit{construction-vehicle}, \textit{barrier} or \textit{traffic-cone} are not annotated for LOD in ParisLuco3D, and are thus not evaluated. Similarly, classes such as our \textit{scootercyclist} are not evaluated, since they are not present in nuScenes.

\subsection{Scans for qualitative evaluation}

\begin{figure*}[h!]
    \centering
    \captionsetup[subfigure]{justification=centering}
        \subfloat[Scan 1800. Considered standard]{
            \includegraphics[width=0.45\linewidth]{img/1800.png}
        }
        \hspace{2em}
        \subfloat[Scan 3000. Considered standard]{
            \includegraphics[width=.45\linewidth]{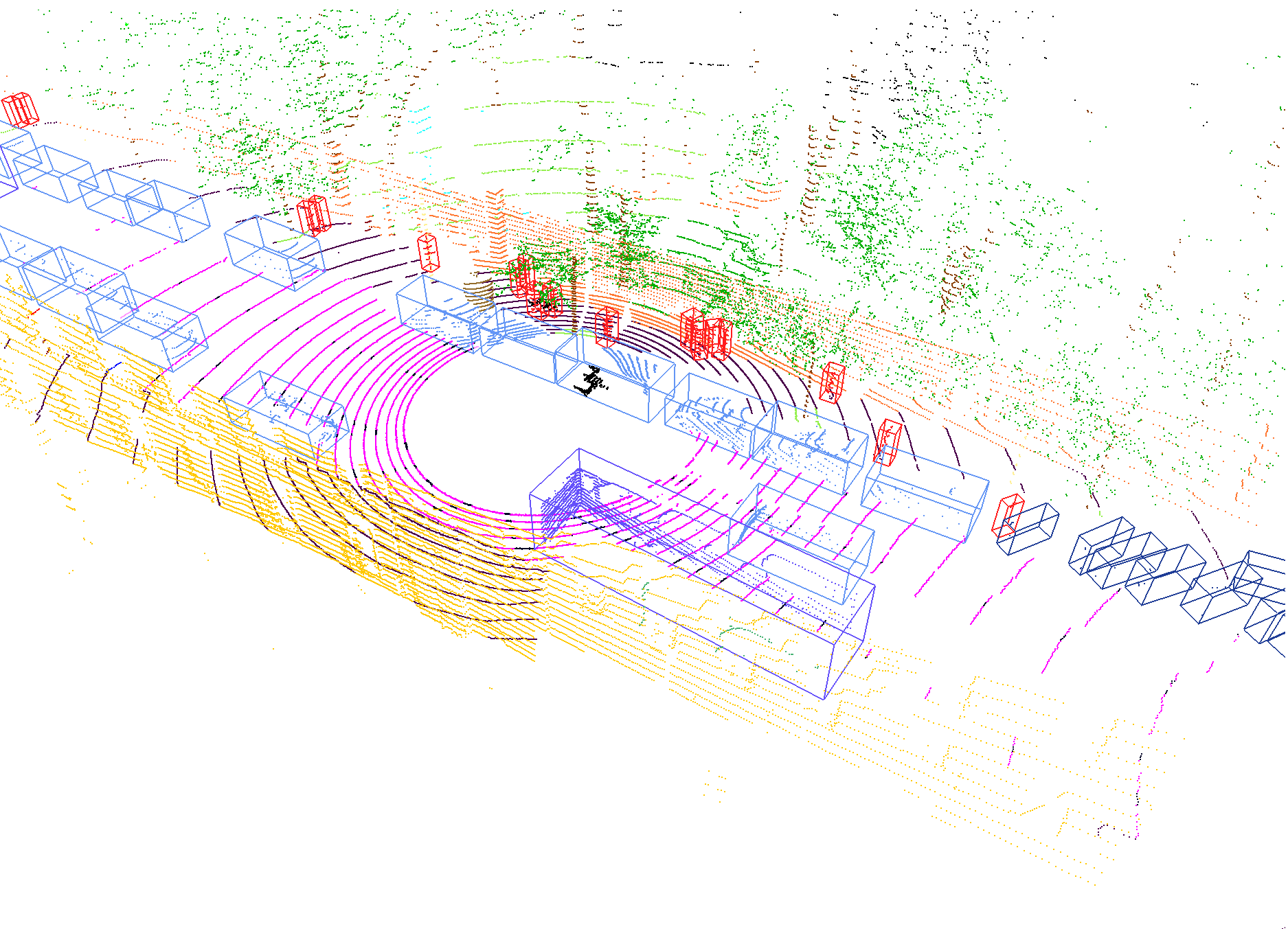}
        }
        \hspace{2em}
        \subfloat[Scan 4500. Considered easy, simple street with few road users and no fence]{
            \includegraphics[width=.45\linewidth]{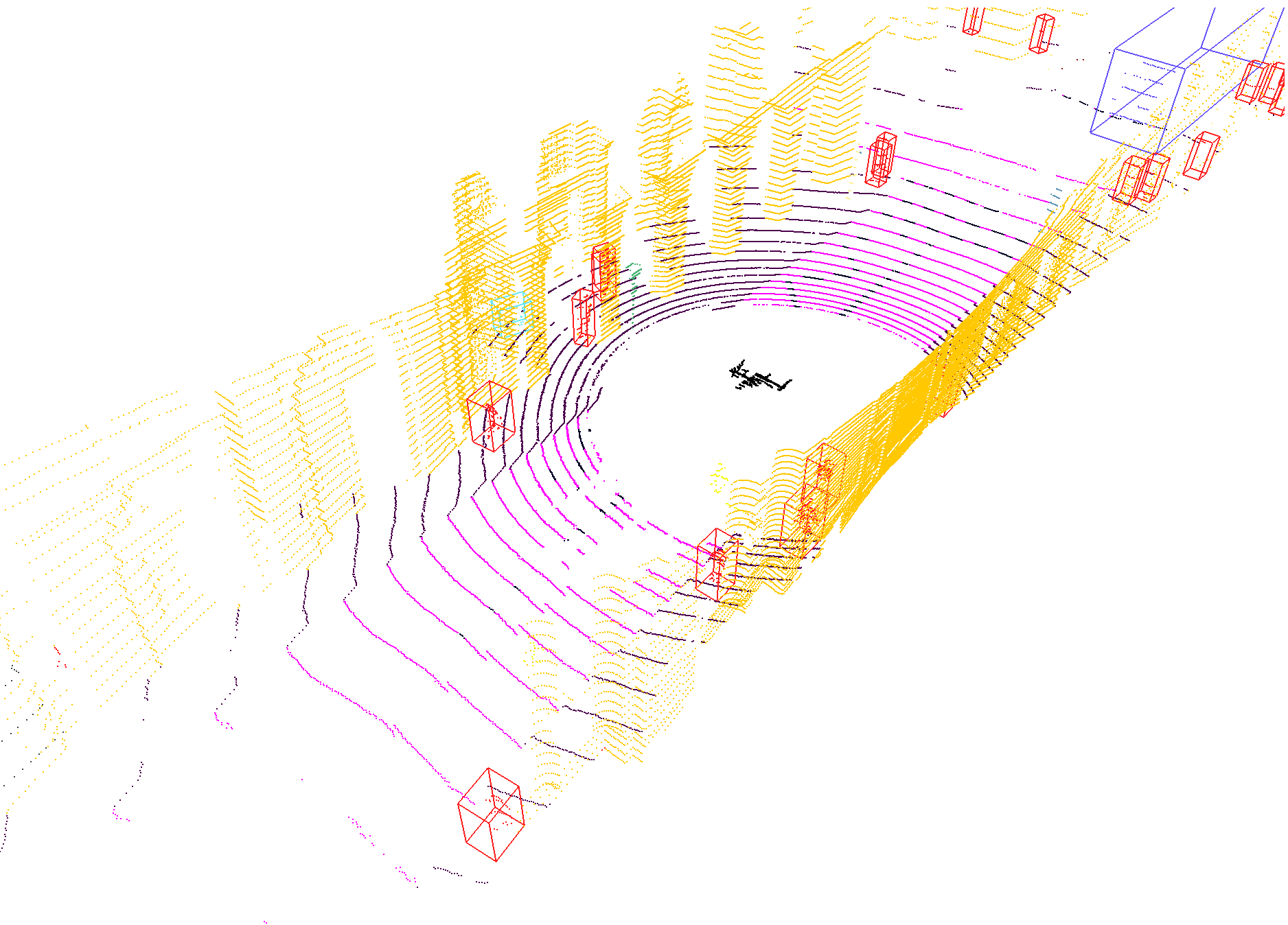}
        }\\
        \subfloat[Scan 5800. Considered hard, near a roundabout under heavy rain]{
            \includegraphics[width=.45\linewidth]{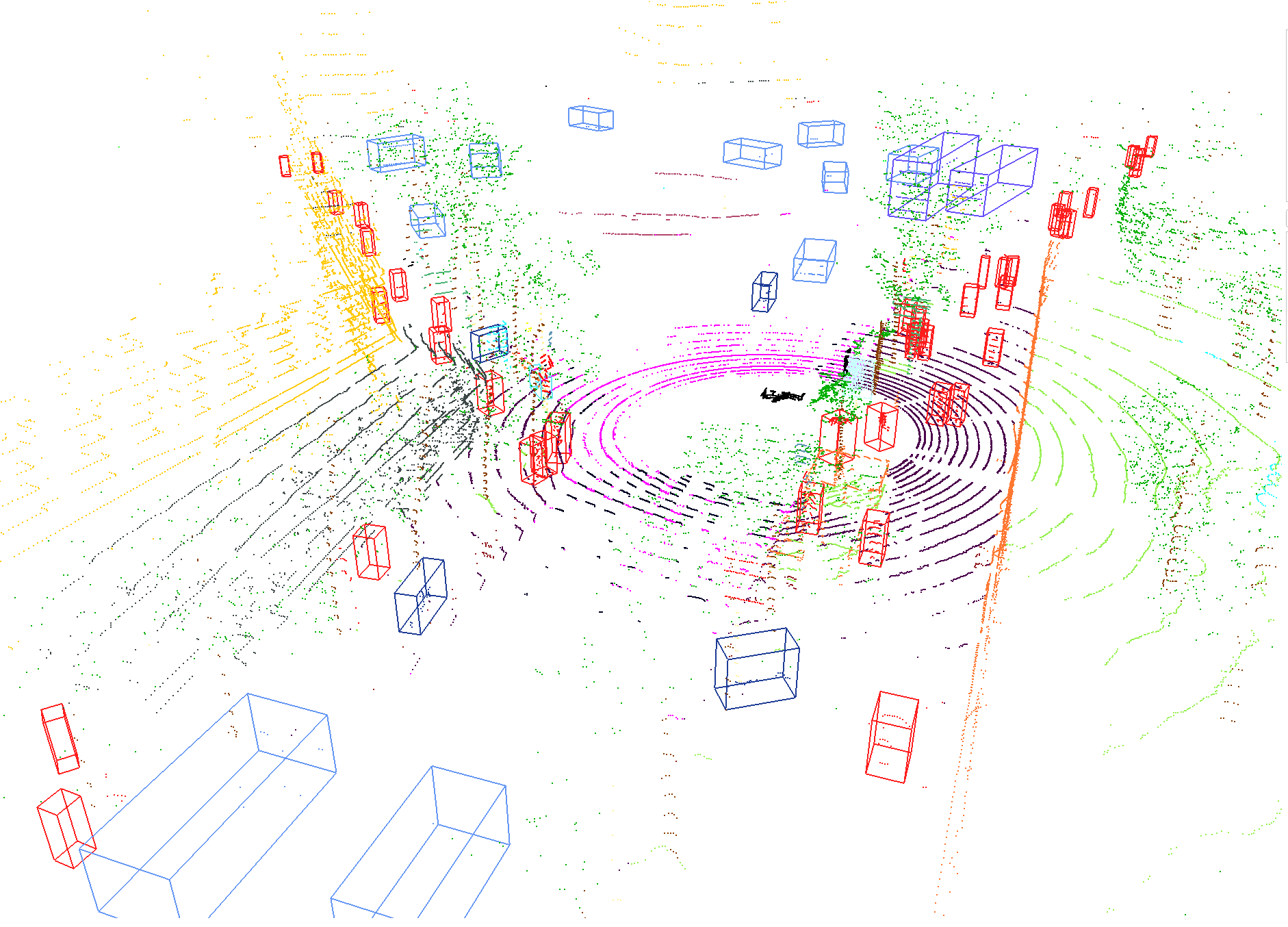}
        }
        \hspace{2em}
        \subfloat[Scan 6500. Considered hard, large amount of road users under heavy rain]{
            \includegraphics[width=.45\linewidth]{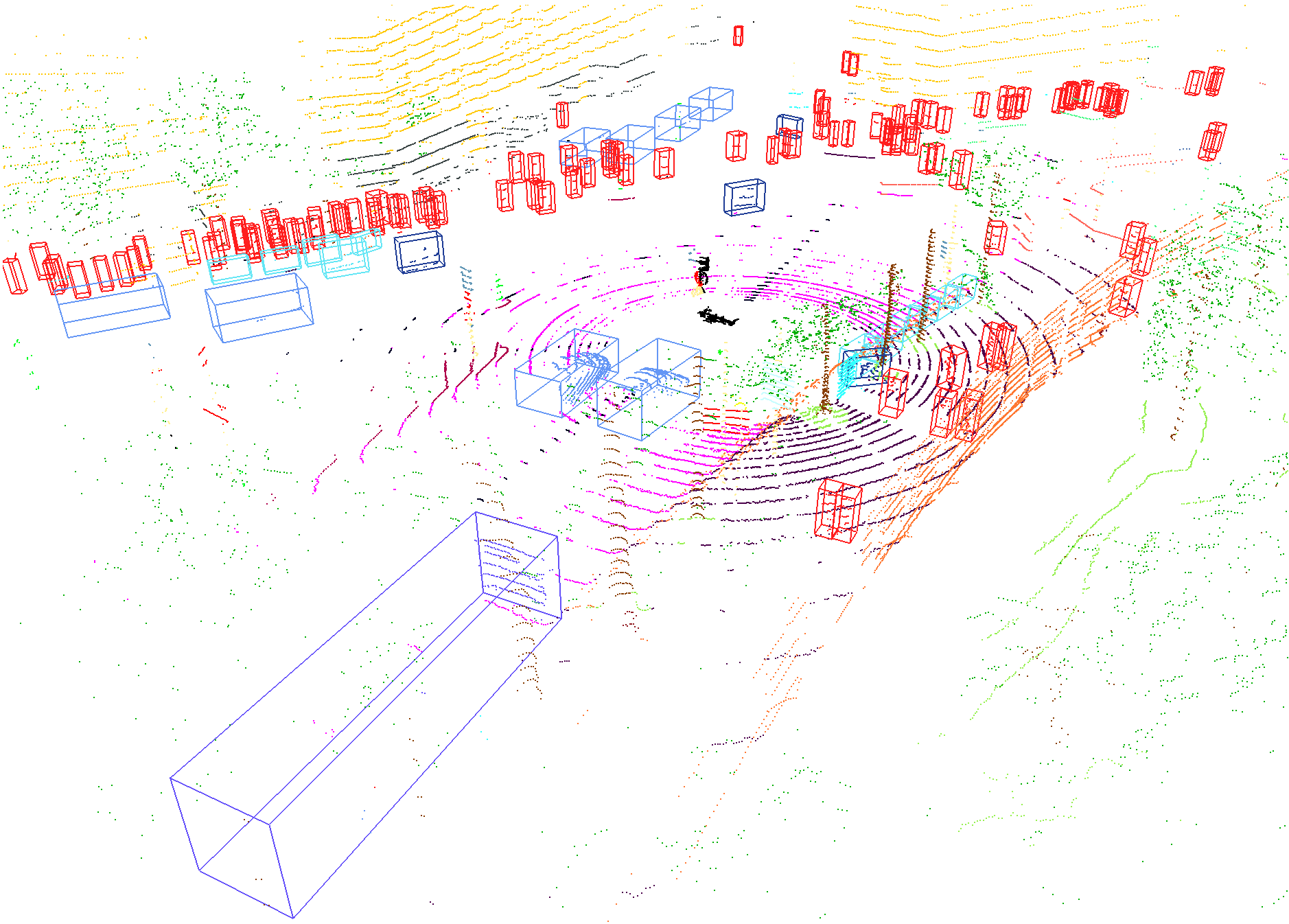}
        }

    \caption{The 5 scans released for qualitative evaluation.}
    \label{fig:quali_frames}
\end{figure*}

Five scans are released with ground truth semantic labels and ground-truth object detection for qualitative comparisons of future methods (shown in \autoref{fig:quali_frames}).

\subsection{Quantitative results}

Quantitative results of our benchmark are in \autoref{tab:SSDGPL3DSK}, \autoref{tab:SSDGPL3DNS}, \autoref{tab:nusc_res_sup}, and \autoref{tab:once_res_sup}.

\begin{table*}
\scriptsize
    \centering
    \begin{tabular}{c|c|c|c|c|c|c|c|c|c|c|c|c|c|c|c|c|c|c|c|c||c}
        & \rotatebox{90}{Car} & \rotatebox{90}{Bicycle} & \rotatebox{90}{Motorcycle}& \rotatebox{90}{Truck}&\rotatebox{90}{Other-vehicle} &\rotatebox{90}{Person} &\rotatebox{90}{Bicyclist}&\rotatebox{90}{Motorcyclist}& \rotatebox{90}{Road}& \rotatebox{90}{Parking}&\rotatebox{90}{Sidewalk} & \rotatebox{90}{Other-ground}&\rotatebox{90}{Building} & \rotatebox{90}{Fence}&\rotatebox{90}{Vegetation} &\rotatebox{90}{Trunk} & \rotatebox{90}{Terrain}& \rotatebox{90}{Pole}& \rotatebox{90}{Traffic-sign}& \rotatebox{90}{mIoU} & \rotatebox{90}{mIoU$_{SK}$} \\ \hline 
        CENet&27.1&0.0&0.3&1.2&19.9&5.8&2.5&2.2&64.3&2.3&56.2&2.8&64.9&2333&44.0&28.1&6.9&35.0&24.5&21.6&58.8\\\hline
        PolarSeg&0.4&0.0&0.0&0.0&7.1&0.0&0.0&0.0&31.2&0.7&24.7&0.4&46.4&21.9&22.4&6.0&11.3&8.8&1.0&9.6&61.8\\\hline
        KPConv&39.7&5.8&8.9&0.3&5.1&30.8&8.8&0.3&8.1&0.6&41.5&0.7&58.6&11.7&67.0&49.9&14.1&26.8&7.1&20.3&61.8\\\hline
        SRUNet&70.5&0.3&12.4&4.7&32.1&20.6&22.6&1.3&71.5&0.1&66.8&0.1&67.3&18.0&71.5&44.9&15.8&41.8&19.3&30.7&63.2\\\hline
        SPVCNN&67.5&0.4&12.3&4.5&18.9&21.9&17.2&0.0&66.7&0.2&67.2&0.1&66.3&12.0&71.7&44.2&11.0&39.5&27.2&28.9&63.4\\\hline
        Helix4D&15.8&0.5&0.0&0.1&4.7&3.0&2.3&0.4&55.1&2.7&42.8&2.1&40.5&22.2&51.1&27.4&7.7&28.8&38.1&18.3&63.9\\\hline
        Cylinder3D&47.1&2.7&2.5&0.3&15.2&11.5&6.8&0.0&58.9&3.9&57.3&1.7&65.9&36.7&54.7&24.8&10.8&32.6&3.6&23.0&64.9\\\hline \hline
       
       \tabcell{SRUNet\\+RayDrop}&78.7&1.1&13.6&7.4&46.9&39.7&31.5&1.6&75.8&1.8&72.6&0.1&71.7&22.9&68.2&54.7&16.8&48.0&27.3&35.8&61.7\\ \hline

       \tabcell{SRUNet\\+IBN-Net}&63.6&4.2&6.1&2.9&30.3&18.6&16.6&3.2&68.3&0.6&59.4&4.4&69.6&29.5&72.6&17.6&13.3&32.7&17.9&28.0&64.9\\

    \end{tabular}
    \caption{Generalization baseline when trained on SemanticKITTI and evaluated on ParisLuco3D. mIoU$_{SK}$ denotes the result on the validation split of SemanticKITTI.}
    \label{tab:SSDGPL3DSK}
\end{table*}

\begin{table*}
\scriptsize
    \centering
    \begin{tabular}{c|c|c|c|c|c|c|c|c|c|c|c|c|c|c|c|c|c||c}
        & \rotatebox{90}{barrier}& \rotatebox{90}{bicycle}& \rotatebox{90}{bus}& \rotatebox{90}{car}& \rotatebox{90}{cnstrctn-vhcl}& \rotatebox{90}{motorcycle}& \rotatebox{90}{pedestrian}& \rotatebox{90}{traffic-cone}& \rotatebox{90}{trailer}& \rotatebox{90}{truck}& \rotatebox{90}{drvbl-grnd}& \rotatebox{90}{other-flat}& \rotatebox{90}{sidewalk}&\rotatebox{90}{terrain} & \rotatebox{90}{manmade}& \rotatebox{90}{vegetation}& \rotatebox{90}{mIoU}& \rotatebox{90}{mIoU$_{NS}$} \\ \hline
        CENet&4.1&4.7&35.7&61.6&1.6&22.7&51.6&0&0&6.1&77.4&22.5&56.7&13.3&81.1&66.6&31.6&69.1 \\ \hline
        PolarSeg&2.1&0.3&1.4&3.2&0.5&1.1&7.0&0&0&0.8&28.7&0.4&20.5&14.0&50.9&49.5&11.3& 71.4 \\ \hline
        KPConv&2.6&0.3&15.9&39.8&0.3&12.0&44.5&0.2&0&2.0&56.2&0.1&17.5&11.7&82.8&79.8&22.9& 64.2 \\ \hline
        SRUNet&4.1&6.4&61.6&82.6&6.5&35.2&60.9&1.2&0&18.8&68.6&7.8&51.2&22.6&87.7&83.6&37.4&69.3  \\ \hline
        SPVCNN&6.5&4.5&64.5&80.5&4.1&36.2&63.2&0.9&0&15.0&71.1&16.5&61.7&22.8&89.2&82.7&38.7&66.8  \\ \hline
        Helix4D&1.0&0.4&9.2&29.2&0.1&2.4&9.3&0&0&0.1&57.4&5.8&40.8&13.0&71.5&65.3&19.2& 69.3 \\ \hline
        Cylinder3D&0.1&0.1&51.9&65.6&1.0&18.8&3.2&1.7&0&9.3&71.&8.4&61.3&1.6&74.4&38.8&25.5& 74.8 \\ \hline \hline
        
       \tabcell{SRUNet\\+RayDrop}&1.5&0.1&62.5&61.8&3.4&21.0&55.6&1.0&0&12.7&68.0&1.4&35.0&17.4&81.2&84.0&31.6&66.4  \\ \hline

       \tabcell{SRUNet\\+IBN-Net}&8.8&9.0&69.1&82.0&5.0&35.9&59.9&1.3&0&18.7&72.8&11.5&60.5&28.7&89.4&82.5&39.7& 67.3\\ 

    \end{tabular}
    \caption{Generalization baseline when trained on nuScenes and evaluated on ParisLuco3D. mIoU$_{NS}$ denotes the result on the validation split of nuScenes.}
    \label{tab:SSDGPL3DNS}
\end{table*}

\begin{table*}
\scriptsize
    \centering
    \begin{tabular}{c|c|c|c|c|c|c|c|c|c|c|c|c||c}
        & \rotatebox{90}{car}& \rotatebox{90}{bus}& \rotatebox{90}{truck}& \rotatebox{90}{trailer}& \rotatebox{90}{bicycle}& \rotatebox{90}{bicyclist}& \rotatebox{90}{motorcycle}& \rotatebox{90}{motorcyclist}& \rotatebox{90}{scooter}& \rotatebox{90}{scootercyclist}& \rotatebox{90}{pedestrian}& \rotatebox{90}{mAP}& \rotatebox{90}{mAP$_{ON}$} \\ \hline
        PointRCNN& 12.6& 5.3& 5.2& -& \multicolumn{4}{c|}{2.8} & \multicolumn{2}{c|}{-} & 15.7&8.3&28.6\\ \hline
        PointPillars& 2.6& 4.4& 1.6& -& \multicolumn{4}{c|}{1.1} & \multicolumn{2}{c|}{-} & 52.3&12.4&45.5\\ \hline
        PV-RCNN& 6.5& 7.8& 3.2& -& \multicolumn{4}{c|}{2.1} & \multicolumn{2}{c|}{-} & 33.6&10.6&52.4\\ \hline
        SECOND& 5.3& 6.5& 1.6& -& \multicolumn{4}{c|}{1.9} & \multicolumn{2}{c|}{-} & 34.1&9.9&54.0\\ \hline
        CenterPoint& 4.3 & 4.1& 1.8& - & \multicolumn{4}{c|}{1.7} & \multicolumn{2}{c|}{-} & 30.6&8.5&59.5\\ \hline
        \change{VoxelNeXt}& \change{3.2} & \change{3.8}& \change{0.8}& - & \multicolumn{4}{c|}{\change{1.5}} & \multicolumn{2}{c|}{-} & \change{16.6}&\change{5.2}&\change{32.2}\\ \hline
        \change{DSVT}& \change{7.0} & \change{5.7}& \change{6.9}& - & \multicolumn{4}{c|}{\change{2.8}} & \multicolumn{2}{c|}{-} & \change{61.9}&\change{16.9}&\change{64.4}\\ \hline\hline
        \tabcell{PointPillars\\+RayDrop}& 4.7& 3.9 & 4.9& -& \multicolumn{4}{c|}{2.1} & \multicolumn{2}{c|}{-} & 51&13.3&39.4\\ \hline
        \tabcell{PointPillars\\+IBN-Net}& 2.7& 3.1& 1.2& -& \multicolumn{4}{c|}{1.4} & \multicolumn{2}{c|}{-} & 47.2&11.1&45.0\\

    \end{tabular}
    \caption{Generalization baseline for 3D Object Detection when trained on ONCE and evaluated on ParisLuco3D. We evaluate on the intersection of classes between the two datasets, following our mapping procedure. This results in 5 classes used for evaluation.}
    \label{tab:once_res_sup}
\end{table*}    

\begin{table*}
\scriptsize
    \centering
    \begin{tabular}{c|c|c|c|c|c|c|c|c|c|c|c|c||c}
        & \rotatebox{90}{car}& \rotatebox{90}{bus}& \rotatebox{90}{truck}& \rotatebox{90}{trailer}& \rotatebox{90}{bicycle}& \rotatebox{90}{bicyclist}& \rotatebox{90}{motorcycle}& \rotatebox{90}{motorcyclist}& \rotatebox{90}{scooter}& \rotatebox{90}{scootercyclist}& \rotatebox{90}{pedestrian}& \rotatebox{90}{mAP}& \rotatebox{90}{mAP$_{NS}$} \\ \hline
        PointRCNN& 25.2&6.7& 5.0& 0.0 & \multicolumn{2}{c|}{0.9}& \multicolumn{2}{c|}{1.6} & \multicolumn{2}{c|}{-} & 34.9&10.6&18.4\\ \hline
        PointPillars& 15.1& 4.0& 1.4& 0.0& \multicolumn{2}{c|}{1.0}& \multicolumn{2}{c|}{1.1} & \multicolumn{2}{c|}{-} & 53.4&10.9&35.3\\ \hline
        PV-RCNN& 29.5& 15.7& 15.8& 0.0& \multicolumn{2}{c|}{0.4}& \multicolumn{2}{c|}{1.8} & \multicolumn{2}{c|}{-} & 58.1&17.3&29.6\\ \hline
        SECOND& 22.2& 3.8& 7.0& 0.0& \multicolumn{2}{c|}{0.5}& \multicolumn{2}{c|}{1.6} & \multicolumn{2}{c|}{-} & 55.1&12.9&38.6\\ \hline
        CenterPoint& 19.0& 6.1& 8.2& 0.0 & \multicolumn{2}{c|}{1.5}& \multicolumn{2}{c|}{3.4} & \multicolumn{2}{c|}{-} & 55.3&13.4&35.0\\ \hline
        \change{VoxelNeXt}& \change{9.2}& \change{0.0}& \change{1.9}& \change{0.0} & \multicolumn{2}{c|}{\change{2.2}}& \multicolumn{2}{c|}{\change{2.2}} & \multicolumn{2}{c|}{-} & \change{62.2}&\change{11.1}&\change{42.2}\\ \hline
        \change{DSVT}& \change{3.1}& \change{0.5}& \change{1.0}& \change{0.0} & \multicolumn{2}{c|}{\change{0.2}}& \multicolumn{2}{c|}{\change{0.3}} & \multicolumn{2}{c|}{-} & \change{58.9}&\change{9.1}&\change{44.2}\\ \hline\hline
        \tabcell{PointPillars\\+RayDrop}& 10.9& 2.2 & 2.8& 0.0& \multicolumn{2}{c|}{0.4}& \multicolumn{2}{c|}{1.5} & \multicolumn{2}{c|}{-} & 48.5&9.5&29.8\\ \hline
        \tabcell{PointPillars\\+IBN-Net}& 8.9& 2.0& 4.0& 0.0& \multicolumn{2}{c|}{0.3}& \multicolumn{2}{c|}{1.7} & \multicolumn{2}{c|}{-} & 55.4&10.3&35.9\\

    \end{tabular}
    \caption{Generalization baseline for 3D Object Detection when trained on nuScenes and evaluated on ParisLuco3D. We evaluate on the intersection of classes between the two datasets, following our mapping procedure. This results in 7 classes used for evaluation.}
    \label{tab:nusc_res_sup}
\end{table*}    


\end{document}